\documentclass[journal]{IEEEtran}

\usepackage{amsmath,graphicx}
\usepackage{algorithm}
\usepackage{algorithmic}
\usepackage{psfrag}

\usepackage{bm}
\usepackage{amssymb,mathtools}
\usepackage{amsfonts}
\usepackage{amsmath,amssymb}
\usepackage{amsthm}
\usepackage{cite}
\usepackage[dvipsnames]{xcolor}
\usepackage{url}
\usepackage[caption=false,subrefformat=parens,labelformat=parens]{subfig}
\usepackage[letterpaper, left=0.625in, right=0.625in, bottom=1in, top=0.75in]{geometry}
\usepackage[hyperfootnotes=false,hidelinks]{hyperref}
\usepackage{graphicx}
\usepackage{upgreek}
\usepackage{yfonts}
\usepackage{cleveref}
\usepackage{amsmath}
\usepackage{arydshln}
\usepackage{cite}
\usepackage{mathtools}
\usepackage{multirow}

\DeclareMathAlphabet\mathbfcal{OMS}{cmsy}{b}{n}
\newcommand{\E}{\mathbb{E}}
\usepackage{balance}

\newcommand{\wbf}{\mathbf{w}}
\newcommand{\xbf}{\mathbf{x}}
\newcommand{\ybf}{\mathbf{y}}

\newcommand{\Abf}{\mathbf{A}}
\newcommand{\Bbf}{\mathbf{B}}
\newcommand{\Cbf}{\mathbf{C}}

\newcommand{\Ibf}{\mathbf{I}}

\newcommand{\Sbf}{\mathbf{S}}

\newcommand{\Xbf}{\mathbf{X}}

\newcommand{\Bcal}{\mathcal{B}}
\newcommand{\Ucal}{\mathcal{U}}

\newcommand{\Ecal}{\mathcal{E}}

\newcommand{\Scal}{\mathcal{S}}

\newcommand{\Acal}{\mathcal{A}}

\newcommand{\Fcal}{\mathcal{F}}

\newcommand{\Jcal}{\mathcal{J}}

\newcommand{\Qcal}{\mathcal{Q}}

\newcommand{\Rcalbf}{\boldsymbol{\mathcal{R}}}
\newcommand{\Hcalbf}{\boldsymbol{\mathcal{H}}}

\newcommand{\Ccalbf}{\boldsymbol{\mathcal{C}}}
\newcommand{\Dcalbf}{\boldsymbol{\mathcal{D}}}
\newcommand{\Scalbf}{\boldsymbol{\mathcal{S}}}

\newcommand{\Kcalbf}{\boldsymbol{\mathcal{K}}}

\newcommand{\Ebb}{\mathbb{E}}

\usepackage{url}

\usepackage{graphicx}

\begin{document}

\title{Resilience in Online Federated Learning: Mitigating Model-Poisoning Attacks via Partial Sharing}
%
\author{Ehsan Lari, \IEEEmembership{Student Member, IEEE}, Reza Arablouei, Vinay Chakravarthi Gogineni, \IEEEmembership{Senior Member, IEEE}, and Stefan Werner, \IEEEmembership{Fellow, IEEE}
\thanks{
This work was supported by the Research Council of Norway. Part of this work has been presented to the 2024 IEEE International Conference on Acoustics, Speech and Signal Processing (ICASSP), Seoul, South Korea, Apr. 2024 [DOI: 10.1109/ICASSP48485.2024.10447497]. \textit{(Corresponding author: Ehsan Lari.)}}
\thanks{Ehsan Lari and Stefan Werner are with the Department of Electronic Systems, Norwegian University of Science and Technology, Trondheim 7491, Norway (e-mail: \{ehsan.lari, stefan.werner\}@ntnu.no). Stefan Werner is also with the Department of Information and Communications Engineering, Aalto University, 00076, Finland.}
\thanks{Reza Arablouei is with the Commonwealth Scientific and Industrial Research Organisation, Pullenvale, QLD 4069, Australia (e-mail: reza.arablouei@csiro.au).}
\thanks{Vinay Chakravarthi Gogineni is with the SDU Applied AI and Data Science, The Maersk Mc-Kinney Moller Institute, Faculty of Engineering, University of Southern Denmark, Odense 5230, Denmark (e-mail: vigo@mmmi.sdu.dk).}
}

\markboth{IEEE TRANSACTIONS ON SIGNAL AND INFORMATION PROCESSING OVER NETWORKS, VOL. XX, NO. XX, AUGUST 2024}
{Shell \MakeLowercase{\textit{et al.}}: Bare Demo of IEEEtran.cls for IEEE Journals}

\maketitle

\begin{abstract}


\textcolor{black}{Federated learning (FL) allows training machine learning models on distributed data without compromising privacy. However, FL is vulnerable to model-poisoning attacks where malicious clients tamper with their local models to manipulate the global model. In this work, we investigate the resilience of the partial-sharing online FL (PSO-Fed) algorithm against such attacks. PSO-Fed reduces communication overhead by allowing clients to share only a fraction of their model updates with the server. We demonstrate that this partial sharing mechanism has the added advantage of enhancing PSO-Fed's robustness to model-poisoning attacks. Through theoretical analysis, we show that PSO-Fed maintains convergence even under Byzantine attacks, where malicious clients inject noise into their updates. Furthermore, we derive a formula for PSO-Fed's mean square error, considering factors like stepsize, attack probability, and the number of malicious clients. Interestingly, we find a non-trivial optimal stepsize that maximizes PSO-Fed's resistance to these attacks. Extensive numerical experiments confirm our theoretical findings and showcase PSO-Fed's superior performance against model-poisoning attacks compared to other leading FL algorithms.}

\end{abstract}

\begin{IEEEkeywords}
Online federated learning, Byzantine clients, partial-sharing, linear regression, model-poisoning.
\end{IEEEkeywords}

\section{Introduction} \label{sec:intro}

\IEEEPARstart{F}{ederated} learning (FL) \cite{mcmahan2017communication,smith2017federated} is a distributed learning paradigm that enables a network of devices, such as smartphones, internet of things (IoT) nodes, and media streamers, to jointly learn a global model without the need to directly disclose their local raw data to the server or other devices. FL is particularly beneficial when data is distributed across numerous devices, and conducting centralized training by transferring the data to the server is unfeasible or impractical \cite{9062302,9170559}.
FL exhibits two key features: the ability to handle heterogeneous data, such as non-independent and identically distributed (non-IID) or unbalanced data, and the ability to run on heterogeneous devices that may have limited resources \cite{yang2019federated,9528995,sspehsan}. 

The communication load in FL refers to the volume of data transmitted between the central server and participating devices during training. It can affect the efficiency and scalability of FL in real-world applications as excessive data transmission may escalate resource usage and the cost associated with training~\cite{9933811,9746228,apsipaehsan}. Several algorithms have been proposed to alleviate the communication load in FL.
Among them, sign stochastic gradient descent (SignSGD)~\cite{bernstein2018signsgd,jin2020stochastic} utilizes sign-based gradient compressors to lower the communication and computational costs of FL. Compressive sensing FL (CS-Fed)~\cite{csfed} reduces the communication load using $1$-bit compressive sensing in analog aggregation transmissions. Quantile sketch FL (QS-Fed)~\cite{fetchsgd} improves communication efficiency by compressing model updates using a count sketch. Online-Fed \cite{9689337} enhances communication efficiency by engaging only a randomly selected subset of clients to communicate with the server in each iteration. However, the aforementioned works do not consider the potential presence of adversaries within the network who may seek to undermine the learning process. This can be through tampering with message exchanges or intentionally providing corrupted or incorrect information. Factoring in the possibility of malicious clients in the network is an important aspect of developing reliable and robust FL algorithms.

In network environments, Byzantine clients are those that act unpredictably or with malicious intent. Such clients can disrupt normal operations by sending misleading information or not following established protocols. This behavior poses significant risks to the FL process, potentially leading to disruption or degraded performance~\cite{guerraoui2018hidden,9857216}. 
Byzantine clients can jeopardize the integrity and reliability of the global model in FL. They may engage in deceptive practices, such as submitting false or inconsistent gradients or model weights during updates, or even initiating denial-of-service (DoS) attacks, all of which can compromise the integrity and effectiveness of the FL process~\cite{chen2017distributed,bernstein2019signsgd,fang2020local}.
To mitigate the impact of Byzantine clients in FL, various strategies have been proposed. These include the use of robust aggregate statistics, the assignment of trust scores to individual clients, and the utilization of historical information to recover from Byzantine attacks~\cite{cao2020fltrust,cao2022fedrecover}.

Byzantine clients may launch a wide range of adversarial attacks such as data-poisoning, model-poisoning, and label-poisoning~\cite{10024252,10113678}. Data-poisoning attacks involve the injection of malicious data into the training dataset \cite{9618642,9855872}. Evasion attacks involve manipulating input data to undermine the generalizability of the learned model while evading detection~\cite{biggio2013evasion}. Backdoor attacks involve embedding hidden trigger mechanisms within the model that can induce targeted malicious behavior~\cite{liu2018trojaning}. Inference-time attacks, such as membership inference, aim to uncover whether a specific data point has been used in training, raising privacy concerns \cite{7958568,9793586}.

Attacks can also target network infrastructure, such as distributed DoS attacks and man-in-the-middle attacks, disrupting services or intercepting communications~\cite{douligeris2007network}. Model-poisoning attacks, where Byzantine clients manipulate model updates during training, are prominent examples~\cite{9771619}. Defense strategies against model-poisoning attacks in FL are reviewed in~\cite{9771619}, categorized into evaluation methods for local model updates and aggregation methods for the global model. In evaluation methods, the server examines client submissions without having access to local client data, which poses challenges to their applicability. In aggregation methods, the server adjusts the local model updates based on certain criteria or utilizes relevant statistical methods. Robust aggregation techniques using median or trimmed mean can mitigate the impact of malicious clients on the global model \cite{yin2018byzantine}. Blockchain-assisted aggregation schemes can also help cope with network attacks but often incur high communication load or computational complexity~\cite{10129156}.

A SignSGD-based communication-efficient FL algorithm has recently been proposed in \cite{10398739} that exhibits resilience to Byzantine attacks. However, it assumes the consistent participation of all clients in every FL iteration, which may not align with the practical realities of FL, particularly in scenarios where clients (such as mobile or IoT devices) have limited resources in terms of energy or computational capacity. Therefore, there is a notable gap in the current FL research: the need for communication-efficient FL algorithms that are robust against model-poisoning attacks without imposing additional computational burden on clients. Addressing this gap can make FL more accessible and practical for a wider range of devices, especially those with constrained resources, and enhance the robustness and scalability of FL systems in real-world applications.

\textcolor{black}{In this paper, we study the impact of partial parameter sharing in online FL~\cite{9746228,9933811,icasspehsan} on the resilience against Byzantine attacks by malicious clients. Our primary motivation stems from a key observation: partial sharing, initially introduced to enhance communication efficiency in online FL, offers a surprising additional benefit by mitigating the negative effects of model-poisoning attacks. Remarkably, this advantage is achieved without imposing any additional computational burden on participating clients.
In summary, our main contributions in this work are:
\begin{itemize}
\item We analyze PSO-Fed under Byzantine attacks (malicious clients tampering with local models) and reveal that partial sharing, designed for communication efficiency, also mitigates model-poisoning attacks.
\item We introduce an intermittent model-poisoning attack model along with random scheduling of the clients within the FL framework.
\item Our rigorous analysis demonstrates that PSO-Fed convergence even with model-poisoning attacks, given an appropriate stepsize.
\item We derive the theoretical MSE of PSO-Fed and validate its accuracy through extensive simulations.
\item We identify a non-trivial optimal stepsize for PSO-Fed under poisoning attacks, contrasting with other algorithms where smaller steps are always better. This finding highlights unique optimization considerations for robust FL.
\item Our thorough experiments show PSO-Fed exhibits greater resilience to model-poisoning attacks compared to existing algorithms, without any additional computational overhead for participating clients.
\end{itemize}}

We organize the remainder of the paper as follows. In section \ref{sec:prealgo}, we provide an overview of the system model, PSO-Fed algorithm, and model-poisoning attacks. In section \ref{sec:analysis}, we evaluate the robustness of PSO-Fed against Byzantine attacks by analyzing its theoretical mean and mean-square convergence. In this section, we also calculate PSO-Fed's theoretical steady-state MSE and optimal stepsize. In section \ref{sec:simulations}, we verify our theoretical findings through a series of numerical experiments. In these experiments, we evaluate the performance of PSO-Fed under model-poisoning attacks through both theoretical predictions and numerical simulations. \textcolor{black}{In section \ref{sec:discussion}, we provide a more detailed explanation of our contributions and discuss the potential benefits and impacts of our work.} Finally, in section \ref{sec:conclusion}, we present some concluding remarks.

\section{\textcolor{black}{Problem Formulation}} \label{sec:prealgo}

In this section, we first describe the considered system model. We then briefly outline the PSO-Fed algorithm~\cite{9746228}, which is a communication-efficient variant of the Online-Fed algorithm~\cite{9689337}. Afterward, we define the model-poisoning Byzantine attack within the context of FL and probe its impact on the model update equations of PSO-Fed.

\subsection{System Model}

We consider a network consisting of $K$ clients and a central server. At every time instance $n$, each client $k$ has access to a data vector $\xbf_{k, n} \in \mathbb{R}^D$ and its corresponding response value $y_{k, n} \in \mathbb{R}$, which are related via the model
\begin{align}\label{eq1}
    y_{k, n} = \wbf^\intercal \xbf_{k, n} + \nu_{k, n},
\end{align}
where the model parameter vector $\wbf \in \mathbb{R}^D$ is collaboratively estimated using the locally stored client data, and $\nu_{k, n}$ is the observation noise. We define the global objective function for estimating $\wbf$ as
\begin{align}\label{eq2}
    \Jcal(\wbf) = \frac{1}{K} \sum\limits_{k=1}^{K} \Jcal_k(\wbf)
\end{align}
with the local objective function at client $k$ being
\begin{align}\label{eq3}
    \Jcal_k(\wbf) &= \E\left[|y_{k, n} -\wbf^\intercal \xbf_{k, n}|^2 \right].
\end{align}
The goal is to find the optimal estimate of $\wbf$ by minimizing $\Jcal(\wbf)$, i.e., $ \wbf^{\star}= \arg\min_{{\wbf}}^{} \Jcal(\wbf)$, albeit, in a distributed fashion via FL.

\subsection{PSO-Fed}

In PSO-Fed, to reduce the communication load, the server only sends a portion of the global model estimate to the clients. Similarly, the clients transmit parts of their local model estimates to the server.
The model parameters exchanged between client $k$ and the server at iteration $n$ are specified using a diagonal selection matrix with $M$ ones, which is denoted by ${\bf S}_{k, n}\in\mathbb{R}^{D \times D}$. The positions of the ones on the diagonal determine which model parameters are shared with the server at each iteration. They can be selected arbitrarily or in a round-robin fashion as in \cite{6671443,6822582} such that the model parameters are exchanged between each client and the server, on average, $M$ times in every $D$ iterations. Therefore, the probability of any model parameter being shared with the server in any iteration is $p_e = M/D$.

Using the selection matrices ${\bf S}_{k, n}$, the recursive update equations of PSO-Fed, that iteratively minimize \eqref{eq2} through FL, are expressed as \cite{9933811}
\begin{subequations}
\begin{align} \label{PSO-Fed}
\epsilon_{k, n}&= y_{k, n}- \left[ \Sbf_{k, n} \wbf_{n} +  \left( \Ibf_D -\Sbf_{k, n} \right) \wbf_{k,n} \right]^\intercal \xbf_{k, n} \\
\wbf_{k, n+1}&= \Sbf_{k, n} \wbf_{n} +  \left( \Ibf_D -\Sbf_{k,n} \right) \wbf_{k,n} + \mu \hspace{.2mm} \xbf_{k, n}  \epsilon_{k, n}\\
\wbf_{n+1}&= \frac{1}{|\Scal_n|} \sum_{k \in \Scal_n }\left[ \Sbf_{k, n+1}  {\wbf}_{k, n+1} + \left( \Ibf_D -\Sbf_{k, n+1} \right) \wbf_{n} \right], \label{psoagg}
\end{align}
\end{subequations}
where $\wbf_{k, n}$ is the local model estimate at client $k$ and iteration $n$, $\wbf_{n}$ is the global model estimate at iteration $n$, $\Ibf_D$ is the $D \times D$ identity matrix, $\mu$ is the stepsize controlling the convergence rate and steady-state performance, $\Scal_n$ denotes the set of client selected at iteration $n$, and $|\Scal_n|$ denotes the number of the selected clients in each iteration.

\subsection{Model-Poisoning Attacks}

We denote the set of potential Byzantine clients within the network as $\Scal_B$. To represent the presence or absence of Byzantine behavior in a client, we use the indicator variable $\beta_k$, where $\beta_k = 1$ indicates that client $k$ belongs to the set $\Scal_B$ (i.e., is a Byzantine client), and $\beta_k = 0$ otherwise. The total number of Byzantine clients in the network is denoted by $|\Scal_B|$. These Byzantine clients seek to poison the global model (i.e., undermine its accuracy) by sporadically sending corrupt local model estimates to the server. We assume that the Byzantine clients possess the accurate local model estimates.
More specifically, we consider a scenario where, in each iteration, every Byzantine client deliberately corrupts its local model estimate by perturbing it, before sending it to the server. This corruption is conducted with a certain probability of attack, denoted by $p_a$. Consequently, we represent the model update shared by any Byzantine client as~\cite{7563348}
\begin{equation}
  {\wbf}^{\prime}_{k, n} =
    \begin{cases}
      \wbf_{k,n} + \boldsymbol{\delta}_{k,n} & \text{with probability $p_a$}\\
      \wbf_{k,n} & \text{with probability $1-p_a$},
    \end{cases} 
\end{equation}
where $\boldsymbol\delta_{k,n} \in \mathbb{R}^D$ denotes the perturbation signal associated with the attack. Typically, this perturbation is zero-mean white Gaussian noise, characterized as $\boldsymbol\delta_{k,n} \sim \mathcal{N}(\mathbf{0},\sigma_B^2 \mathbf{I}_D)$ \cite{7563348}.

The probability of a Byzantine client corrupting its local estimate can be modeled using a Bernoulli random variable $\tau_{k,n}$. This variable takes the value of $1$ with probability $p_a$ indicating the occurrence of an attack, and $0$ otherwise. Hence, given that each Byzantine client transmits a corrupted local model estimate to the server with a probability of $p_a$ when selected by the server, in scenarios where Byzantine clients are present in the network, the global model update of PSO-Fed, \eqref{psoagg}, can be rewritten as
 \begin{equation}\label{PSO-Fed_byz}
 \wbf_{n+1} = \frac{1}{|\Scal_n|} \sum_{k \in \Scal_n }\left[ \Sbf_{k, n+1}  {\wbf}^{\prime}_{k, n+1} + \left( \Ibf_D -\Sbf_{k, n+1} \right) \wbf_{n} \right]\notag
 \end{equation}
where
\begin{equation}\label{eq:attack_model}
{\wbf}^{\prime}_{k, n+1} = \wbf_{k,n+1} + \beta_k\tau_{k,n} \boldsymbol \delta_{k,n+1}.
\end{equation}

\section{Performance Analysis} \label{sec:analysis}

In this section, we evaluate the robustness of PSO-Fed against Byzantine attacks by analyzing its theoretical mean and mean-square convergence and predicting its steady-state MSE and optimal stepsize. 

To facilitate the analysis, we introduce some new quantities. We denote the extended optimal global model as ${\bf w}^{\star}_{e} = {\bf 1}_{K+1} \otimes {\bf w}^{\star}$ and the extended global model estimate as ${\bf w}_{e, n}=\mathrm{col}\{{\bf w}_{n}, \wbf_{1, n}, \ldots, \wbf_{K,n}\}$. We also define the following collective quantities
\begin{align*}
{\Xbf}_n&=\mathrm{bdiag}\{\textbf{0}, {\xbf}_{1, n}, \ldots, {\xbf}_{K, n}\}\\
\boldsymbol \delta_{e, n}&=\mathrm{col}\big\{ \textbf{0}, \boldsymbol \delta_{1, n}, \ldots, \boldsymbol \delta_{K, n} \big\}\\
\mathbf{B}&=\mathrm{bdiag}\big\{ 0, \beta_{1}\Ibf_D, \ldots, \beta_{K}\Ibf_D \big\}\\
\mathbf{T}_{n}&=\mathrm{bdiag}\big\{ 0, \tau_{1,n}\Ibf_D, \ldots, \tau_{K,n}\Ibf_D\big\}\\
\boldsymbol \nu_{e, n}&=\mathrm{col}\big\{{0},\nu_{1, n}, \ldots,\nu_{K, n} \big\},
\end{align*}
where the operators $\mathrm{col}\{\cdot\}$ and $\mathrm{bdiag}\{\cdot\}$ represent column-wise stacking and block diagonalization, respectively. Additionally, ${\bf 1}_{K+1}$ denotes a column vector with $K+1$ entries, all set to one.
Subsequently, we define
\begin{subequations} \label{y_e_col}
\begin{align}
{\bf y}_{e, n}&= \mathrm{col}\{0, y_{1, n}, y_{2, n}, \ldots, y_{K, n}\}={\Xbf}^{\intercal}_{n} {\bf w}_{e}^{\star} + \boldsymbol{\nu}_{e, n} \\
\boldsymbol{\epsilon}_{e, n}&=\mathrm{col}\big\{ 0, \epsilon_{1, n}, \epsilon_{2, n}, \ldots, \epsilon_{K, n} \big\}= {\bf y}_{e, n} - {\Xbf}^{\intercal}_{n} \boldsymbol{\Acal}_{n} {\bf w}_{e, n}
\end{align}
\end{subequations}
\begin{align}\label{eq9}
&\boldsymbol{\Acal}_{n} = \\
&\begin{bmatrix}
    \Ibf_{D} & \textbf{0} & \textbf{0} & \dots  & \textbf{0} \\
    a_{1, n} \Sbf_{1, n} & \Ibf_{D} - a_{1, n} \Sbf_{1, n} & \textbf{0} & \dots  & \textbf{0} \\
    \vdots & \vdots & \vdots & \ddots & \vdots \\
    a_{K, n} \Sbf_{K, n} & \textbf{0} & \textbf{0} & \dots  & \Ibf_{D} - a_{K, n} \Sbf_{K, n}  \notag
\end{bmatrix},
\end{align}
where $a_{k, n}=1$ if $k\in\Scal_n$ and $a_{k, n}=0$ otherwise. Hence, the global recursion equations of PSO-Fed can be expressed as
\begin{align}\label{eq10}
{\bf w}_{e, n+1}= \boldsymbol{\mathcal{B}}_{n+1} \big(\boldsymbol{\mathcal{A}}_{n} {\bf w}_{e, n}+ \mu{\Xbf}_n\boldsymbol{\epsilon}_{e, n} \big) + {\Ccalbf}_{n+1} \mathbf{B}\mathbf{T}_{n} \boldsymbol \delta_{e, n+1}
\end{align}
where
\begin{align}\label{eq11}
&\boldsymbol{\Bcal}_{n+1} =  \\
&\begin{bmatrix}
    \Ibf_{D} - \hspace{-1mm} \sum_{k=1}^{K} \hspace{-1mm} \frac{a_{k, n}}{|\Scal_n|}  \Sbf_{k, n+1}& \frac{a_{1, n}}{|\Scal_n|}  \Sbf_{1, n+1}  & \dots  & \frac{a_{K, n}}{|\Scal_n|}  \Sbf_{K, n+1} \\
    \textbf{0} & \Ibf_{D}  & \dots  & \textbf{0} \\
    \vdots & \vdots  & \ddots & \vdots \\
    \textbf{0} & \textbf{0} & \dots  & \Ibf_{D} \notag
\end{bmatrix}
\end{align}
and
\begin{align}
{\Ccalbf}_{n+1} = 
& \begin{bmatrix}
    \textbf{0} & \frac{a_{1, n}}{|\Scal_n|}  \Sbf_{1, n+1}  & \dots  & \frac{a_{K, n}}{|\Scal_n|}  \Sbf_{K, n+1} \\
    \textbf{0} &  \textbf{0} & \dots  & \textbf{0} \\
    \vdots & \vdots  & \ddots & \vdots \\
    \textbf{0} & \textbf{0} & \dots  & \textbf{0}
\end{bmatrix}. 
\end{align}

To make the analysis tractable, we adopt the following commonly used assumptions.

\noindent A1: The input vectors of each client, ${\xbf}_{k,n}$, originate from a wide-sense stationary (WSS) multivariate random process characterized by the covariance matrix ${\bf R}_{k} = \Ebb [{\xbf}_{k,n} {\xbf}_{k,n}^{\intercal}]$.

\noindent A2: The observation noise, $\nu_{k,n}$, and perturbation signal, $\boldsymbol\delta_{k, n}$, both follow identical and independent distributions. They are also independent of all other stochastic variables.

\noindent A3: The selection matrices, $\Sbf_{k,n}$, of all clients and iterations are independent of each other.


\subsection{Mean Convergence}

Let us define the deviation (coefficient-error) vector as $\widetilde{\mathbf{w}}_{e,n}={\bf w}_{e}^{\star}-{\bf w}_{e,n}$. Since $\boldsymbol{\Bcal}_{n+1}$ and $\boldsymbol{\Acal}_{n}$ are block right-stochastic, i.e., their block rows add up to identity matrix, we have $\boldsymbol{\Bcal}_{n+1} {\bf w}_{e}^{\star} = \boldsymbol{\Acal}_{n} {\bf w}_{e}^{\star} = {\bf w}_{e}^{\star}$. Therefore, considering \eqref{eq10} and the definition of $\widetilde{\mathbf{w}}_{e,n}$, we get
\begin{align} \label{eq12}
\widetilde{{\bf w}}_{e, n+1}=& \boldsymbol{\Bcal}_{n+1} \big(\mathbf{I} -\mu {\Xbf}_n {\Xbf}_n^\intercal \big) \boldsymbol{\Acal}_{n} \widetilde{{\bf w}}_{e, n}-\mu \boldsymbol{\Bcal}_{n+1} {\Xbf}_n \boldsymbol{\nu}_{e,n} \notag\\-& {\Ccalbf}_{n+1} \mathbf{B} \mathbf{T}_{n} \boldsymbol \delta_{e, n+1}.
\end{align}
Taking the expected value of both sides of \eqref{eq12} and under assumptions A1-A3, we obtain
\begin{equation}\label{eq13}
\Ebb [\widetilde{{\bf w}}_{e, n+1}] = \Ebb[\boldsymbol{\Bcal}_{n+1}] \big(\mathbf{I} -\mu \Rcalbf \big) \Ebb [\boldsymbol{\Acal}_{n}]  \Ebb [\widetilde{{\bf w}}_{e, n}],
\end{equation}
where $$\Rcalbf = \mathrm{bdiag} \{{\bf0}, {\bf R}_{1}, {\bf R}_{2}, \cdots, {\bf R}_{K} \}.$$

The mean convergence of PSO-Fed is guaranteed if $\|\left(\mathbf{I} -\mu \Rcalbf \right)\| < 1$ or equivalently $\|1 -\mu \lambda_i \left({\bf R}_{k}\right)\|<1$ $\forall k,i$. Here, $\|\cdot\|$ is any matrix norm and $\lambda_i \left( \cdot \right)$ denotes the $i$th eigenvalue of its matrix argument. Consequently, PSO-Fed converges in the mean sense, even amidst the presence of Byzantine clients, as long as the stepsize $\mu$ is appropriately chosen to satisfy
\begin{equation}\label{cond}
0 < \mu < \frac{2}{\max_{i,k}\{ \lambda_i \left( {\bf R}_{k} \right) \}}.
\end{equation}
In other terms, PSO-Fed is unbiased, even under model-poisoning attacks.

\subsection{Mean-Square Convergence}

Let us denote the weighted norm-square of $\widetilde{{\bf w}}_{e, n}$ as $\|\widetilde{{\bf w}}_{e, n}\|_{\boldsymbol\Sigma}^2 = \widetilde{{\bf w}}^{\intercal}_{e, n} {\boldsymbol\Sigma} \widetilde{{\bf w}}_{e, n}$, where $\boldsymbol\Sigma$ is a positive semi-definite matrix. Computing the weighted norm-square of both sides of \eqref{eq12} results in
 \begin{align}\label{varrel}
&\E[\|\widetilde{{\bf w}}_{e, n+1}\|_{\boldsymbol \Sigma}^2]\\
&=\E[\|\widetilde{{\bf w}}_{e, n}\|_{\boldsymbol \Sigma^{\prime}}^2] + \mu^2 \E[ \boldsymbol{\nu}^{\intercal}_{e,n} {\bf Y}_n{\boldsymbol \nu}_{e,n}]+\E[ \boldsymbol \delta_{e, n+1}^{\intercal} {\bf U}_n\boldsymbol \delta_{e, n+1}], \notag
\end{align}
where the cross terms vanish under assumption A2 and
\begin{align}\label{sigmaprime} \nonumber
\boldsymbol \Sigma^{\prime}  &= \E[ \boldsymbol{\Acal}^{\intercal}_{n} \big(\mathbf{I} -\mu {\Xbf}_n {\Xbf}_n^\intercal \big) 
\boldsymbol{\Bcal}^{\intercal}_{n+1} \boldsymbol \Sigma \boldsymbol{\Bcal}_{n+1} \big(\mathbf{I} -\mu {\Xbf}_n {\Xbf}_n^\intercal \big) \boldsymbol{\Acal}_{n}]\\ 
{\bf Y}_n &= {\Xbf}_n^\intercal \boldsymbol{\Bcal}^{\intercal}_{n+1} \boldsymbol \Sigma \boldsymbol{\Bcal}_{n+1} {\Xbf}_n\\ \nonumber
{\bf U}_n &= {\mathbf{T}}^\intercal_{n} \mathbf{B}^\intercal {\Ccalbf}^{\intercal}_{n+1} \boldsymbol \Sigma  {\Ccalbf}_{n+1}  \mathbf{B} \mathbf{T}_n.
\end{align}
Subsequently, we define
\begin{align}
    \boldsymbol{\sigma^{\prime}} =&\ \mathrm{bvec}\{\boldsymbol{\Sigma}^{\prime}\} = \boldsymbol \Fcal^{\intercal} \boldsymbol{\sigma}\\
    \boldsymbol{\sigma}=&\ \mathrm{bvec}\{\boldsymbol{\Sigma}\}\\
    {\boldsymbol\Fcal} =&\ {\boldsymbol\Qcal}_{\boldsymbol \Bcal} {\boldsymbol\Qcal}_{\boldsymbol \Acal} - \mu {\boldsymbol\Qcal}_{\boldsymbol \Bcal} \Kcalbf {\boldsymbol\Qcal}_{\boldsymbol \Acal} + \mu^2 {\boldsymbol\Qcal}_{\boldsymbol \Bcal} \Hcalbf {\boldsymbol\Qcal}_{\boldsymbol \Acal} \label{Fmat} \\
    {\boldsymbol\Qcal}_{\boldsymbol \Bcal} =&\ \E[\boldsymbol{\Bcal}_{n+1} \otimes_b \boldsymbol{\Bcal}_{n+1}]\\
    {\boldsymbol\Qcal}_{\boldsymbol \Acal} =&\ \E[{\boldsymbol\Acal}_{n} \otimes_b {\boldsymbol\Acal}_{n}]\\
    \Kcalbf =&\ (\mathbf{I} \otimes_b \Rcalbf) +  (\Rcalbf \otimes_b \mathbf{I})\\
    \Hcalbf =&\ \E[{\Xbf}_n {\Xbf}_n^\intercal \otimes_b {\Xbf}_n {\Xbf}_n^\intercal],
\end{align}
where $\otimes_b$ denotes the block Kronecker product and $\mathrm{bvec}\{\cdot\}$ the block vectorization operation \cite{koning1991block}. We evaluate ${\boldsymbol\Qcal}_{\boldsymbol \Acal}$, ${\boldsymbol\Qcal}_{\boldsymbol \Bcal}$, and $\Hcalbf$, under A1-A3, in Appendixes \ref{Ap2} and \ref{Ap1}. 

The third term on the right-hand side (RHS) of \eqref{varrel} quantifies the impact of model-poisoning attacks. We calculate this term as
\begin{align}
\E&[\boldsymbol\delta_{e, n+1}^{\intercal} {\bf U}_n\boldsymbol\delta_{e, n+1} ]\notag\\
&= \E[ \boldsymbol \delta_{e, n+1}^{\intercal}{\mathbf{T}}^\intercal_{n} \mathbf{B}^\intercal {\Ccalbf}^{\intercal}_{n+1} \boldsymbol \Sigma  {\Ccalbf}_{n+1}\mathbf{B} \mathbf{T}_n\boldsymbol \delta_{e, n+1}]\notag\\
&= \E[\mathrm{tr} ( \boldsymbol \delta_{e, n+1}^{\intercal} { \mathbf{T}}^\intercal_{n} \mathbf{B}^\intercal {\Ccalbf}^{\intercal}_{n+1} \boldsymbol \Sigma {\Ccalbf}_{n+1}  \mathbf{B} \mathbf{T}_n \boldsymbol \delta_{e, n+1} ) ]\notag\\
&= \mathrm{tr} (\E[ \boldsymbol \delta_{e, n+1}^{\intercal} { \mathbf{T}}^\intercal_{n} \mathbf{B}^\intercal {\Ccalbf}^{\intercal}_{n+1} \boldsymbol \Sigma {\Ccalbf}_{n+1}  \mathbf{B} \mathbf{T}_n \boldsymbol \delta_{e, n+1} ] )\notag\\
&= \mathrm{tr} \big(\E[{\Ccalbf}_{n+1}  \E[ \mathbf{B} \mathbf{T}_n \boldsymbol \delta_{e, n+1} \boldsymbol \delta_{e, n+1}^{\intercal} { \mathbf{T}}^\intercal_{n} \mathbf{B}^\intercal ]{\Ccalbf}^{\intercal}_{n+1} ]\boldsymbol \Sigma \big),
\end{align}
where $\mathrm{tr}(\cdot)$ denotes the matrix trace. Under A2, we have
\begin{align}\label{traceofdelta}
\mathrm{tr} & \big(\E[ {\Ccalbf}_{n+1} \E[  \mathbf{B} \mathbf{T}_n \boldsymbol \delta_{e, n+1} \boldsymbol \delta_{e, n+1}^{\intercal}{\mathbf{T}}^\intercal_{n} \mathbf{B}^\intercal  ] {\Ccalbf}^{\intercal}_{n+1} ]\boldsymbol \Sigma \big)\notag\\
& = \mathrm{tr} \big( \E[ {\Ccalbf}_{n+1} \boldsymbol \Omega_{\delta}  {\Ccalbf}^{\intercal}_{n+1} ] \boldsymbol \Sigma  \big),
\end{align}
with 
\begin{align}
\boldsymbol \Omega_{\boldsymbol\delta} &= \E[\mathbf{B} \mathbf{T}_n \boldsymbol\delta_{e,n+1} \boldsymbol\delta^\intercal_{e, n+1}{\mathbf{T}}^\intercal_{n} \mathbf{B}^\intercal]\\
&= \mathbf{B} \E[ \mathbf{T}_n \boldsymbol \delta_{e, n+1}\boldsymbol \delta_{e, n+1}^{\intercal} { \mathbf{T}}_{n} ] \mathbf{B} \notag \\
&= \mathrm{bdiag}\{\mathbf{0}, \beta_{1} \sigma_B^2 p_a \mathbf{I}_D, \beta_{2} \sigma_B^2 p_a \mathbf{I}_D, \cdots , \beta_{K} \sigma_B^2 p_a \mathbf{I}_D\}. \notag 
\end{align}
Using the properties of block Kronecker product, we have $\mathrm{tr} \big( \E[ {\Ccalbf}_{n+1} \boldsymbol \Omega_{\delta} {\Ccalbf}^{\intercal}_{n+1} ]\boldsymbol \Sigma  \big) = \boldsymbol{\omega}^\intercal \boldsymbol{\sigma},$
where 
\begin{align} \label{omegadelta}
\boldsymbol \omega &= \mathrm{bvec}\{ \E[ {\Ccalbf}_{n+1} \boldsymbol \Omega_{\delta} {\Ccalbf}^{\intercal}_{n+1} ]\}\notag\\
& = {\boldsymbol\Qcal}_{\Ccalbf}\mathrm{bvec}\{ \boldsymbol \Omega_{\boldsymbol \delta}\}
\end{align}
and
$${\boldsymbol\Qcal}_{\Ccalbf} = \E[\Ccalbf_{n+1} \otimes_b \Ccalbf_{n+1}]. $$
We evaluate ${\boldsymbol\Qcal}_{\Ccalbf}$ in Appendix \ref{ApQC}.
We also define 
\begin{align}
\boldsymbol{\phi} &= {\boldsymbol\Qcal}_{\boldsymbol \Bcal} \boldsymbol{\phi}_{\boldsymbol \nu}, \notag\\
\boldsymbol{\phi}_{\boldsymbol \nu} &= \mathrm{bvec}\{ \E \left[ {\Xbf}_n {\boldsymbol \Theta}_{\boldsymbol \nu} {\Xbf}_n^\intercal \right] \}~\text{(Appendix \ref{Ap3})}, \notag \\
{\boldsymbol \Theta}_{\boldsymbol \nu} &= \E \left[ {\boldsymbol \nu}_{e,n}{\boldsymbol \nu}^{\intercal}_{e,n}  \right] = \mathrm{diag}\{0, \sigma_{\nu_1}^2, \cdots, \sigma_{\nu_K}^2\}, 
\end{align}
and $\mathrm{bvec}^{-1}\{ \cdot\}$ as the reverse operation of block vectorization. Consequently, we can write the global recursion equation for the weighted mean square deviation (MSD) of PSO-Fed under model-poisoning attacks as
\begin{align}\label{globrec}
\E\left[\|\widetilde{{\bf w}}_{e, n+1}\|_{\mathrm{bvec}^{-1}\{ \boldsymbol \sigma \}}^2\right] =& \ \E\left[\|\widetilde{{\bf w}}_{e, n}\|_{\mathrm{bvec}^{-1}\{\boldsymbol \Fcal^{\intercal} \boldsymbol \sigma \}}^2\right] \notag \\
&+\mu^2 \boldsymbol{\phi}^\intercal  \boldsymbol \sigma + \boldsymbol{\omega}^\intercal \boldsymbol \sigma.
\end{align}

As $n \rightarrow \infty$, \eqref{globrec} converges provided that the spectral radius of $\boldsymbol \Fcal$ is smaller than one, i.e., $\rho({\boldsymbol\Fcal}^\intercal) = \rho({\boldsymbol\Fcal}) < 1$. 
Using the properties of the block maximum-norm, denoted by $\|\cdot\|_{b,\infty}$, and knowing $ \| {\boldsymbol\Qcal}_{\boldsymbol \Bcal}\|_{b,\infty} = \| {\boldsymbol\Qcal}_{\boldsymbol \Acal}\|_{b,\infty} = 1$, we have
\begin{align}\label{SRF}
\rho({\boldsymbol\Fcal}) & \leq \| {\boldsymbol\Qcal}_{\boldsymbol \Bcal} \left( \mathbf{I} - \mu \Kcalbf+ \mu^2 \Hcalbf \right) {\boldsymbol\Qcal}_{\boldsymbol \Acal} \|_{b,\infty} \notag \\ 
&  \leq \|  \mathbf{I} - \mu \Kcalbf+ \mu^2 \Hcalbf  \|_{b,\infty}.
\end{align}
Considering \cite[Theorem 2]{1254028} and defining 
$$\Dcalbf = \begin{bmatrix}
\Kcalbf/2 & -\Hcalbf/2\\ \mathbf{I} & \mathbf{0}
\end{bmatrix},$$
the spectral radius of $\boldsymbol \Fcal$ is smaller than one when
\begin{equation} \label{rhoFcond}
0 < \mu < \min\biggl\{\frac{1}{\lambda_{\max}\left(\Kcalbf^{-1} \Hcalbf \right)},\frac{1}{\max\{ \lambda (\Dcalbf)  \in \mathbb{R}, 0 \}}\biggl\},
\end{equation}
where $\lambda_{\max}(\cdot)$ denotes the largest eigenvalue of its matrix argument. We denote the upper bound of \eqref{rhoFcond} as $\mu_{\max}$. Thus, when \eqref{rhoFcond} holds, PSO-Fed converges in the mean-square sense and has a bounded steady-state MSD, even when facing Byzantine attacks.

\subsection{Mean Square Error} \label{SecMSE}

Utilizing the calculations of the three terms on the RHS of \eqref{varrel} and unfolding the iterations of the global recursion in  \eqref{globrec}, we obtain
\begin{align}\label{varrel2}
&\E\left[\|\widetilde{{\bf w}}_{e, n+1}\|_{\mathrm{bvec}^{-1}\{ \boldsymbol \sigma \}}^2\right] 
=\E\left[\|\widetilde{{\bf w}}_{e, 0}\|_{\mathrm{bvec}^{-1}\{(\boldsymbol \Fcal^{\intercal})^{n+1} \boldsymbol \sigma \}}^2\right]  \notag \\
&+\mu^2 \boldsymbol{\phi}^\intercal \sum_{j=0}^{n} \left(\boldsymbol \Fcal^{\intercal} \right)^j \boldsymbol \sigma + \boldsymbol{\omega}^\intercal \sum_{j=0}^{n} \left( \boldsymbol \Fcal^{\intercal} \right)^j \boldsymbol \sigma.\end{align}  
Given an appropriate choice of $\mu$ as in \eqref{rhoFcond}, letting $n \rightarrow \infty$ on both sides of \eqref{varrel2} yields
\begin{equation}
\lim_{n \to \infty} \E[\|\widetilde{{\bf w}}_{e,n}\|_{\mathrm{bvec}^{-1}\{\boldsymbol \sigma\}}^2] = (\mu^2 \boldsymbol{\phi}^\intercal+\boldsymbol{\omega}^\intercal) \big(\mathbf{I} - \boldsymbol\Fcal^{\intercal}\big)^{-1}\boldsymbol \sigma.
\label{varrel3}
\end{equation}
To calculate the steady-state MSE of PSO-Fed (denoted by $\Ecal$), while taking into account the impact of model-poisoning attacks, we set
\begin{align}
\boldsymbol \sigma&=\mathrm{bvec}\{ \E\left[ \boldsymbol{\Acal}_{n}^{\intercal} {\Xbf}_{n} {\Xbf}^{\intercal}_{n} \boldsymbol{\Acal}_{n} \right] \}\notag \\
&= \E\left[ \mathrm{bvec}\{ \boldsymbol{\Acal}_{n}^{\intercal} {\Xbf}_{n} {\Xbf}^{\intercal}_{n} \boldsymbol{\Acal}_{n} \} \right] \notag \\
&= \E\left[ ({\boldsymbol\Acal}_{n} \otimes_b {\boldsymbol\Acal}_{n})^{\intercal}\mathrm{bvec}\{ {\Xbf}_{n} {\Xbf}^{\intercal}_{n} \} \right]  \notag \\
&= \E\left[ ({\boldsymbol\Acal}_{n} \otimes_b {\boldsymbol\Acal}_{n})^{\intercal} \right] \mathrm{bvec}\{ \E\left[ {\Xbf}_{n} {\Xbf}^{\intercal}_{n} \right] \}  \notag \\
&= {\boldsymbol\Qcal}_{\boldsymbol \Acal}^{\intercal} \mathrm{bvec}\{ \Rcalbf \}.
\label{bvecpsi}
\end{align}
Therefore, considering \eqref{y_e_col}, we have
\begin{align}\label{SSMSE}
\Ecal & = \frac{1}{K} \lim_{n \to \infty} \E[\boldsymbol{\epsilon}_{e, n}^{\intercal} \boldsymbol{\epsilon}_{e, n}] \notag \\ 
& = \frac{1}{K} \lim_{n \to \infty} \left( \E\left[ \widetilde{\mathbf{w}}_{e,n}^{\intercal} \boldsymbol{\Acal}_{n}^{\intercal} {\Xbf}_{n} {\Xbf}^{\intercal}_{n} \boldsymbol{\Acal}_{n} \widetilde{\mathbf{w}}_{e,n} \right] + \E\left[ \boldsymbol{\nu}_{e, n}^{\intercal} \boldsymbol{\nu}_{e, n} \right] \right) \notag \\
& = \frac{1}{K} \left[ (\mu^2 \boldsymbol{\phi}^\intercal+\boldsymbol{\omega}^\intercal) \big(\mathbf{I} - \boldsymbol\Fcal^{\intercal}\big)^{-1} {\boldsymbol\Qcal}_{\boldsymbol \Acal}^{\intercal} \mathrm{bvec}\{ \Rcalbf \} + \mathrm{tr}\left({\boldsymbol \Theta}_{\boldsymbol \nu}\right) \right] \notag \\
& = \underbrace{\frac{\mu^2}{K}\boldsymbol{\phi}^\intercal\big(\mathbf{I} - \boldsymbol\Fcal^{\intercal}\big)^{-1} {\boldsymbol\Qcal}_{\boldsymbol \Acal}^{\intercal} \mathrm{bvec}\{\Rcalbf\}}_{\Ecal_{\boldsymbol{\phi}}}\notag\\
& + \underbrace{\frac{1}{K}\boldsymbol{\omega}^\intercal\big(\mathbf{I} - \boldsymbol\Fcal^{\intercal}\big)^{-1} {\boldsymbol\Qcal}_{\boldsymbol \Acal}^{\intercal} \mathrm{bvec}\{\Rcalbf\}}_{\Ecal_{\boldsymbol{\omega}}} + \underbrace{\frac{1}{K}\mathrm{tr}\left({\boldsymbol \Theta}_{\boldsymbol \nu}\right)}_{\Ecal_{\boldsymbol{\Theta}}}.
\end{align}


The partial sharing of model parameters, scheduling of clients, statistics of input data, and value of stepsize impact the first and second terms on the RHS of~\eqref{SSMSE}, $\Ecal_{\boldsymbol{\phi}}$ and $\Ecal_{\boldsymbol{\omega}}$. The observation noise influences the first and third terms, $\Ecal_{\boldsymbol{\phi}}$ and $\Ecal_{\boldsymbol \Theta}$, with the latter term being solely due to this noise. Moreover, the effect of model-poisoning attacks by Byzantine clients on the steady-state MSE of PSO-Fed is confined to the second term, $\Ecal_{\boldsymbol{\omega}}$, which is induced by these attacks.

\textit{Remark 1:} Considering \eqref{omegadelta} and the calculation of ${\boldsymbol\Qcal}_{\Ccalbf}$ in Appendix \ref{ApQC}, both partial sharing and client scheduling have a diminishing effect on $\boldsymbol\omega$. The lower the probability of sharing each entry of the model parameter vector, $p_e = \frac{M}{D}$, the smaller the entries of $\boldsymbol\omega$ are. Therefore, \eqref{SSMSE} suggests that partial sharing can indeed endow online FL with enhanced resilience to model-poisoning attacks. However, partial sharing also affects $\Ecal_{\boldsymbol{\phi}}$ and increases its value as shown in \cite[Fig. 1(a)]{9933811}. This makes it hard to straightforwardly determine the overall impact of partial sharing on the steady-state MSE of PSO-Fed in scenarios involving model-poisoning attacks, by only analyzing \eqref{SSMSE}. To gain deeper insights and validate these theoretical predictions, we undertake comprehensive simulations in section \ref{sec:simulations}.



\subsection{Optimal Stepsize} \label{OptStep}


In the absence of model-poisoning attacks where $\Ecal_{\boldsymbol{\omega}}=0$, $\Ecal$ is a monotonically increasing function of the stepsize $\mu$, given \eqref{rhoFcond} is satisfied. However, under model-poisoning attacks, the existence of $\Ecal_{\boldsymbol{\omega}}$ disrupts this monotonicity. It amplifies $\Ecal$ for smaller values of $\mu$ and shifts the point of minimal $\Ecal$ from $\mu=0$ to a larger $\mu>0$. Consequently, this leads to the emergence of a non-trivial optimal stepsize that results in the minimum $\Ecal$ when facing model-poisoning attacks. We denote this optimal stepsize as $\mu^*$.

Assuming \eqref{rhoFcond} is satisfied hence $\rho(\boldsymbol\Fcal^{\intercal}) < 1$, approximating $\big(\mathbf{I}-\boldsymbol\Fcal^{\intercal}\big)^{-1}$ with its corresponding $J$-term truncated Neumann series~\cite[Eq. (186)]{IMM201203274} gives
\begin{equation} \label{approxF}
     \big(\mathbf{I} - \boldsymbol\Fcal^{\intercal}\big)^{-1} \approx \sum_{j=0}^{J} \left(\boldsymbol\Fcal^{\intercal} \right)^j.
 \end{equation}
We consider $J\geq3$. Moreover, for simplicity, we rewrite \eqref{Fmat} as 
\begin{align*}
\boldsymbol\Fcal^{\intercal} = \Abf_0 - \mu \Abf_1 + \mu^2 \Abf_2,
\end{align*}
by defining $\Abf_0 = {\boldsymbol\Qcal}_{\boldsymbol \Acal}^{\intercal} {\boldsymbol\Qcal}_{\boldsymbol \Bcal}^{\intercal}$, $\Abf_1 = {\boldsymbol\Qcal}_{\boldsymbol \Acal}^{\intercal} \Kcalbf {\boldsymbol\Qcal}_{\boldsymbol \Bcal}^{\intercal}$, and $\Abf_2 = {\boldsymbol\Qcal}_{\boldsymbol \Acal}^{\intercal} \Hcalbf{\boldsymbol\Qcal}_{\boldsymbol \Bcal}^{\intercal}$.
Using \eqref{approxF} in \eqref{SSMSE}, we have
\begin{align*}
\Ecal \approx  \frac{1}{K} \left[ \left(\mu^2 \boldsymbol{\phi}^\intercal+\boldsymbol{\omega}^\intercal\right) \left\{ \sum_{j=0}^{J} \left(\boldsymbol\Fcal^{\intercal} \right)^j \right\} {\boldsymbol\Qcal}_{\boldsymbol \Acal}^{\intercal} \mathrm{bvec}\{ \Rcalbf \} + \mathrm{tr}\left({\boldsymbol \Theta}_{\boldsymbol \nu}\right) \right].
\end{align*}
Since a valid stepsize is typically much smaller than one, we can further approximate $\mathcal{E}$ by excluding terms beyond $\mu^2$ as 
\begin{align} \label{muopt1}
\Ecal \approx \frac{1}{K} \Big[ \big[& \boldsymbol{\omega}^\intercal \Bbf_{0,J} - \mu\boldsymbol{\omega}^\intercal \Bbf_{1,J} \\
&  + \mu^2 (\boldsymbol{\phi}^\intercal \Bbf_{0,J} + \boldsymbol{\omega}^\intercal \Bbf_{2,J}) \big] {\boldsymbol\Qcal}_{\boldsymbol \Acal}^{\intercal} \mathrm{bvec}\{ \Rcalbf \} + \mathrm{tr}\left({\boldsymbol \Theta}_{\boldsymbol \nu}\right) \Big], \notag 
\end{align}
where
\begin{align*}
\Bbf_{0,J} & = \sum_{j=0}^{J} \Abf_0^j, \ \Bbf_{1,J}  = \sum_{j=1}^{J} \Abf_0^{j-1} \Abf_1 \Bbf_{0,J-j}, \\
\Bbf_{2,J} & = \Bbf_{1,J-1} \Abf_1 + \sum_{j=1}^{J} \Abf_0^{j-1} \Abf_2 \Bbf_{0,J-j} + \sum_{j=3}^{J} \Cbf_{2,J}, \\
\Cbf_{2,J} & = \Cbf_{2,J-1} \Abf_0 + \left( \sum_{j=1}^{J} \Abf_0^{j-1} \Abf_1 \Abf_0^{J-j} \right) \Abf_1, \ \Cbf_{2,2} = \Abf_1^2.
\end{align*}

Differentiating $\mathcal{E}$ in \eqref{muopt1} with respect to $\mu$, setting the derivative equal to zero, and solving for $\mu$ yields the approximate optimal stepsize as
\begin{equation} \label{muopt2}
    \mu^{*} \approx \frac{ \boldsymbol{\omega}^\intercal \Bbf_{1,J} {\boldsymbol\Qcal}_{\boldsymbol \Acal}^{\intercal} \mathrm{bvec}\{ \Rcalbf \} } { 2 (\boldsymbol{\phi}^\intercal \Bbf_{0,J} + \boldsymbol{\omega}^\intercal \Bbf_{2,J}) {\boldsymbol\Qcal}_{\boldsymbol \Acal}^{\intercal} \mathrm{bvec}\{ \Rcalbf \} }.
\end{equation}
Note that, when no model-poisoning attack occurs, i.e., $\boldsymbol{\omega} = 0$, the numerator on the RHS of \eqref{muopt2} becomes zero, leading to $\mu^* = 0$.


\section{Simulation Results} \label{sec:simulations}

To verify our theoretical findings, we conduct several numerical experiments. We also compare the performance of the PSO-Fed algorithm with that of the Online-Fed, SignSGD, CS-Fed, and QS-Fed algorithms within a federated network comprising $ K = 50$ or $100$ clients and a model parameter vector of size $D = 5$. Each client possesses non-IID data vectors $\xbf_{k, n}$ and their associated response values $y_{k, n}$, which are interconnected as per~\eqref{eq1} with model parameter vector $\wbf=\frac{1}{\sqrt{D}}[1,\cdots,1]^\intercal \in \mathbb{R}^{D}$.
Each entry of $\xbf_{k, n}$ for each client $k$ is drawn from a zero-mean Gaussian distribution with variance $\varsigma_{k}^2$, where $\varsigma_{k}^2$ itself is sampled from a uniform distribution between $0.2$ and $1.2$, denoted as $\Ucal(0.2,1.2)$. 
In addition, the observation noise $\nu_{k, n}$ is zero-mean IID Gaussian with variance $\sigma_{\nu_k}^2$ drawn from $\Ucal(0.005, 0.025)$.  In the conducted experiments, only $M = 1$ entry of the model parameter vector is shared per iteration, unless specified differently.

We evaluate the performance of the considered algorithms on the server's side by employing a test dataset consisting of $N_{t}=50$ instances $\{\breve{\Xbf}, \breve{\ybf}\}$ and calculating the test MSE at the server as
\begin{align}
\frac{1}{N_{t}} \ \|\breve{\ybf} - \breve{\Xbf}^{\intercal} \wbf_n\|^2_2. 
\end{align}
We also evaluate the performance of PSO-Fed on the client side using \eqref{PSO-Fed} and calculating the network-wide average steady-state MSE as
\begin{align}
 \frac{1}{K} \ \sum\limits_{k=1}^{K} \lim_{n \rightarrow \infty} \epsilon_{k, n}^2. 
\end{align}

\begin{figure}[t!]
 \centering
 \includegraphics[width=.485\textwidth]{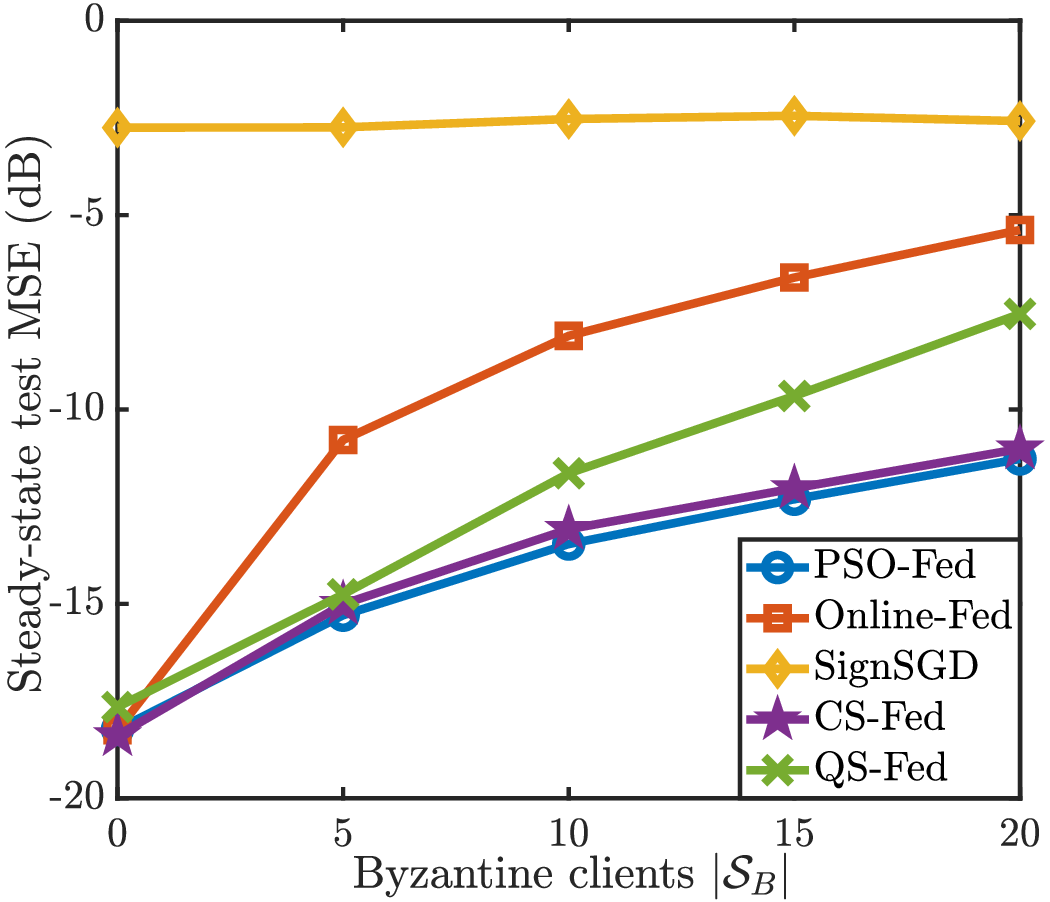}
 \caption{Steady-state test MSE for different algorithms with different numbers of Byzantine clients $|\Scal_B|$, attack strength $\sigma_{B}^2 = 0.25$ and attack probability $p_a = 1$.}
 \label{fig:fig1}
\end{figure}
\begin{figure}[t!]
 \centering
 \includegraphics[width=.485\textwidth]{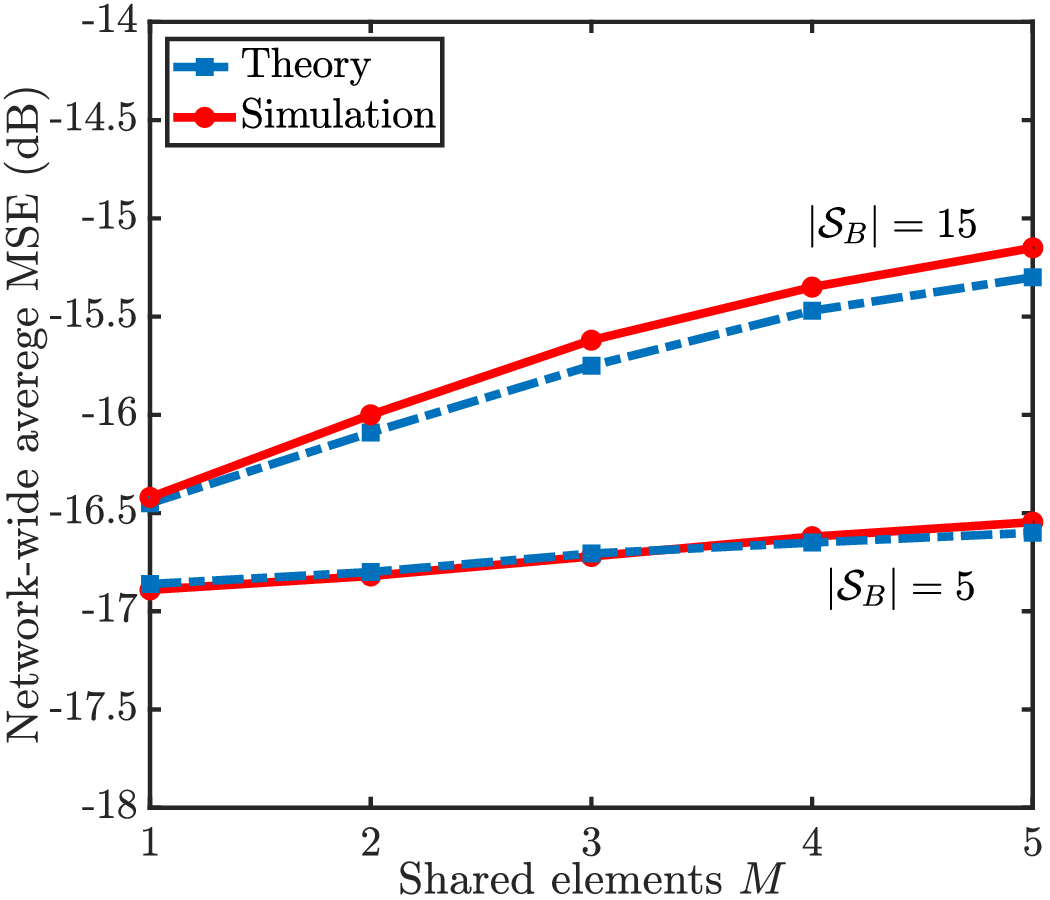}
 \caption{Steady-state test MSE of PSO-Fed for different numbers of shared elements $M$ with different numbers of Byzantine clients $|\Scal_B|$, attack strength $\sigma_{B}^2 = 0.5$ and attack probability $p_a = 0.2$.}
 \label{fig:fig2}
\end{figure}
\begin{figure}[t!]
 \centering
 \includegraphics[width=.485\textwidth]{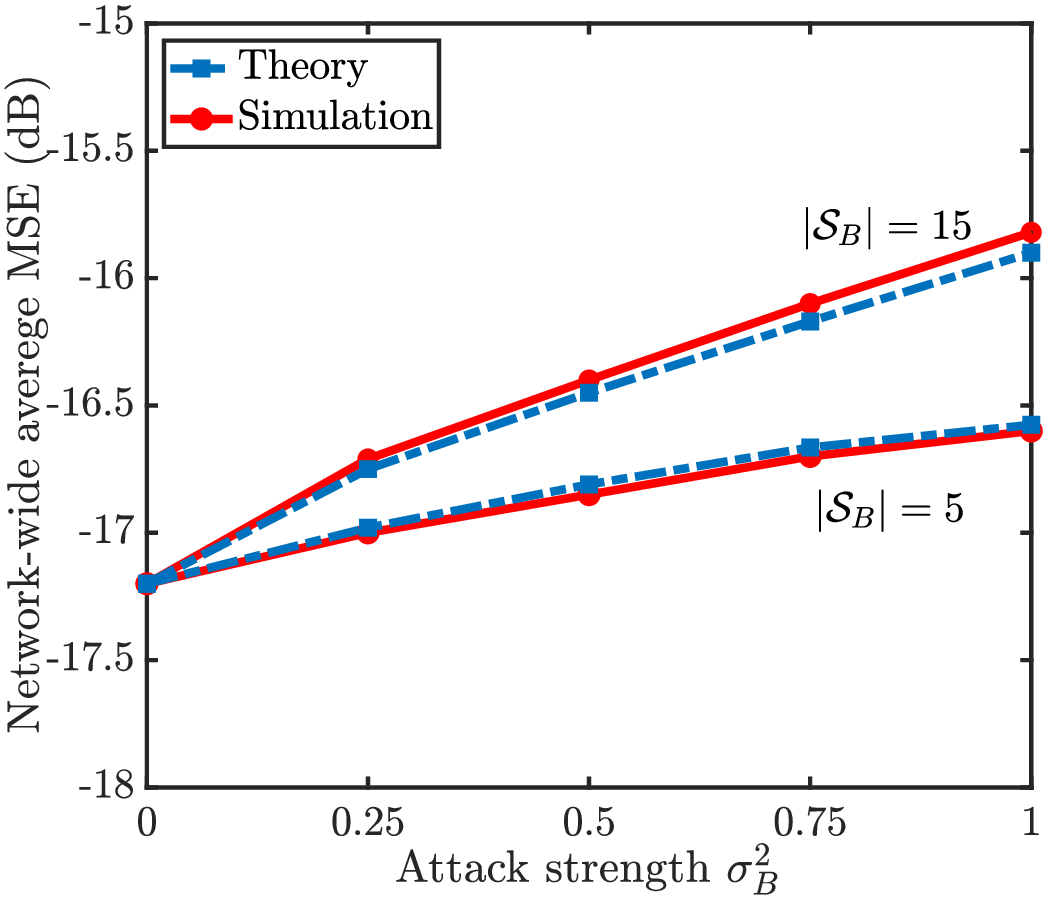}
 \caption{Network-wide average MSE of PSO-Fed for different values of attack strengths $\sigma_{B}^2$ and Byzantine clients $|\Scal_B|$, number of shared elements $M = 1$ and attack probability $p_a = 0.2$.}
 \label{fig:fig3}
\end{figure}
\begin{figure}[t!]
 \centering
 \includegraphics[width=.485\textwidth]{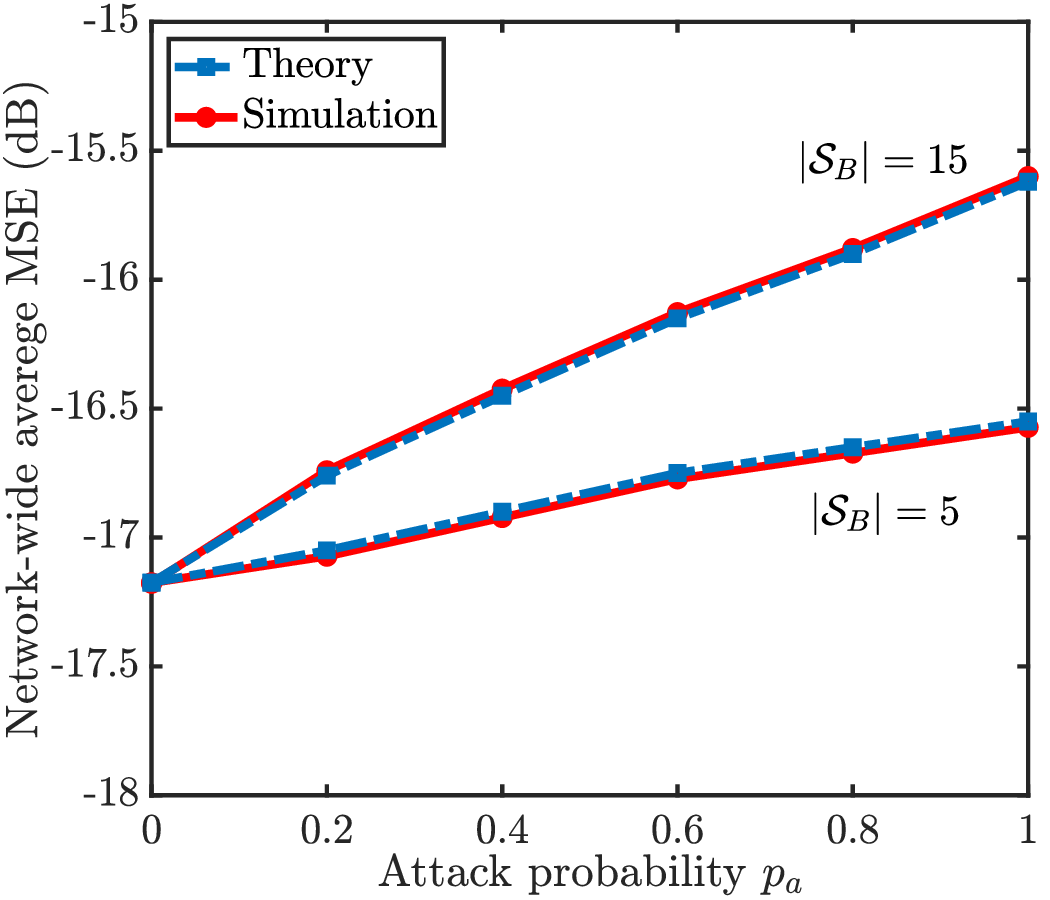}
 \caption{Effect of attack probability $p_a$ on steady-state test MSE of PSO-Fed for different numbers of Byzantine clients $|\Scal_B|$, number of shared elements $M = 1$ and attack strength $\sigma_{B}^2 = 0.25$.}
 \label{fig:fig4}
\end{figure}
\begin{figure}[t!]
 \centering
 \includegraphics[width=.485\textwidth]{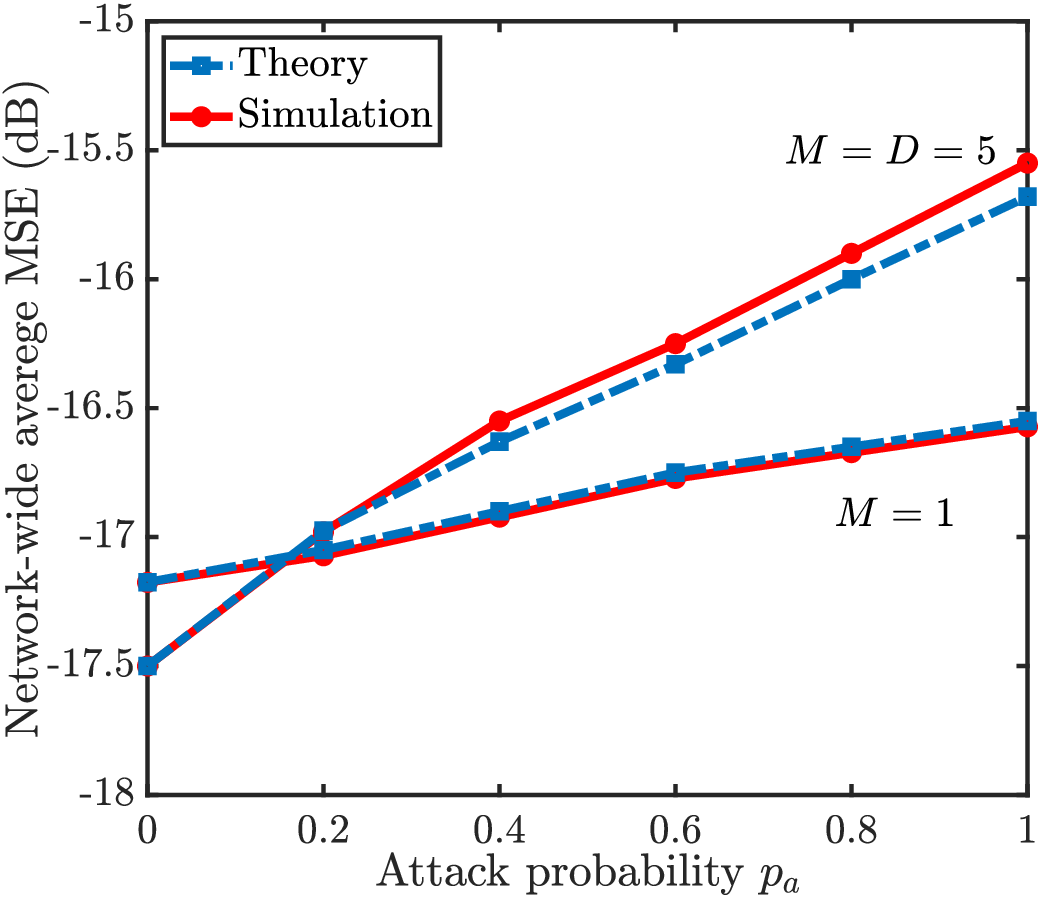}
 \caption{Effect of attack probability $p_a$ on steady-state test MSE of PSO-Fed for different numbers of shared elements $M \in \{1,5\}$, number of Byzantine clients $|\Scal_B| = 5$ and attack strength $\sigma_{B}^2 = 0.25$.}
 \label{fig:fig4_2}
\end{figure}
\begin{figure}[t!]
 \centering
 \includegraphics[width=.485\textwidth]{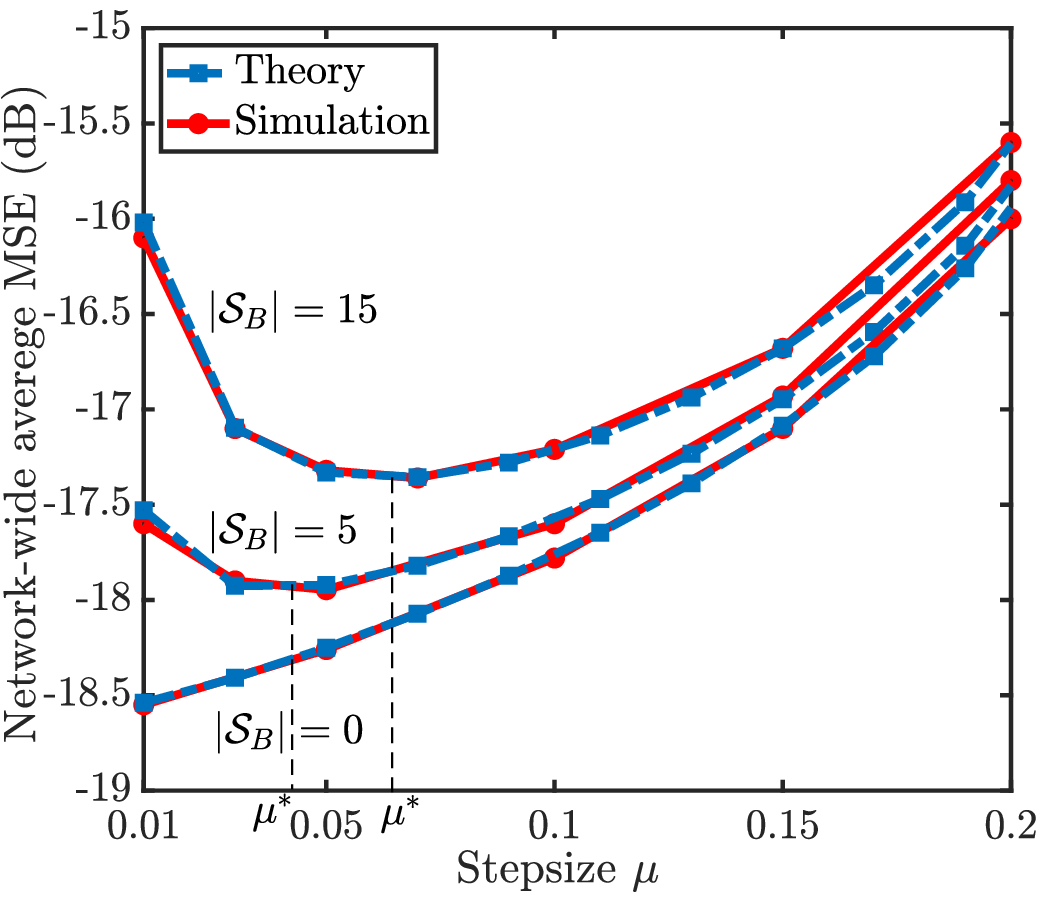}
 \caption{Network-wide average MSE of PSO-Fed for different values of stepsize $\mu$, attack strength $\sigma_{B}^2 = 0.25$ and attack probability $p_a = 0.25$.}
 \label{fig:fig5}
\end{figure}
\begin{figure}[t!]
 \centering
 \includegraphics[width=.485\textwidth]{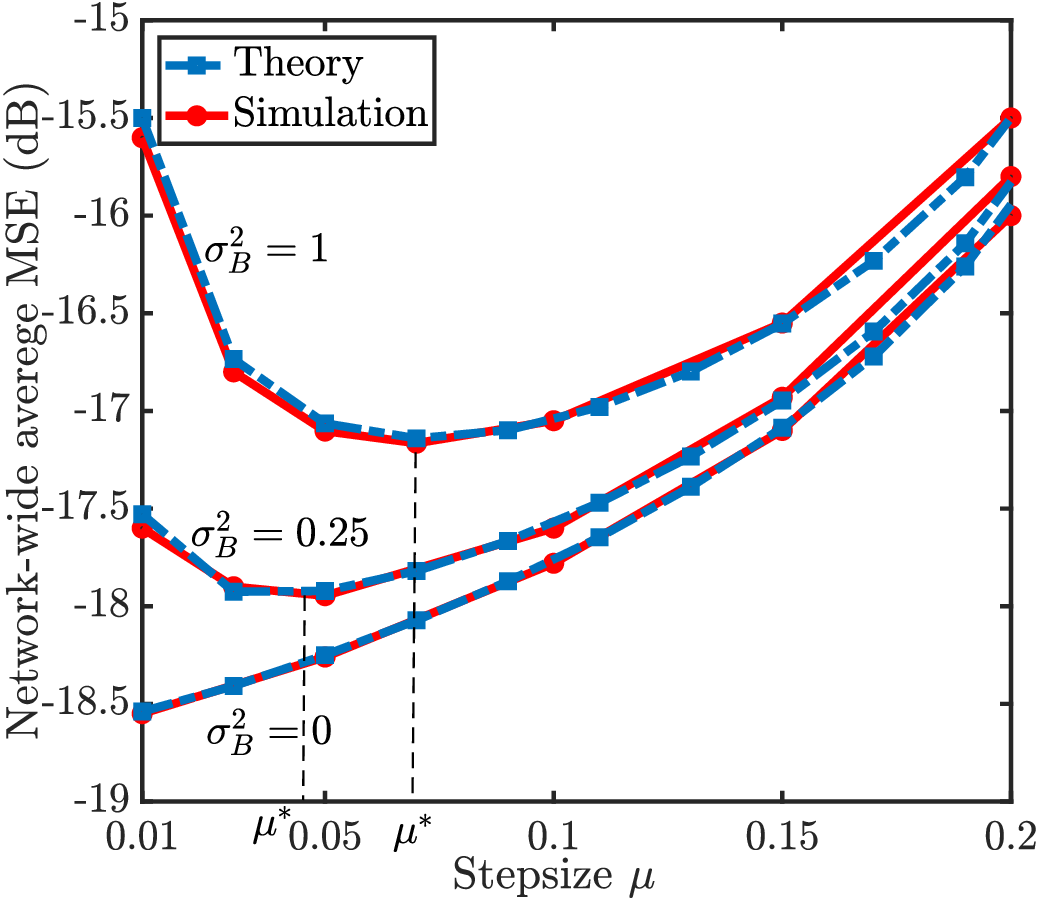}
 \caption{Network-wide average MSE of PSO-Fed for different values of stepsize $\mu$, numbers of Byzantine clients $|\Scal_B | = 5$ and attack probability $p_a = 0.25$.}
 \label{fig:fig6}
\end{figure}
\begin{figure}[t!]
 \centering
 \includegraphics[width=.485\textwidth]{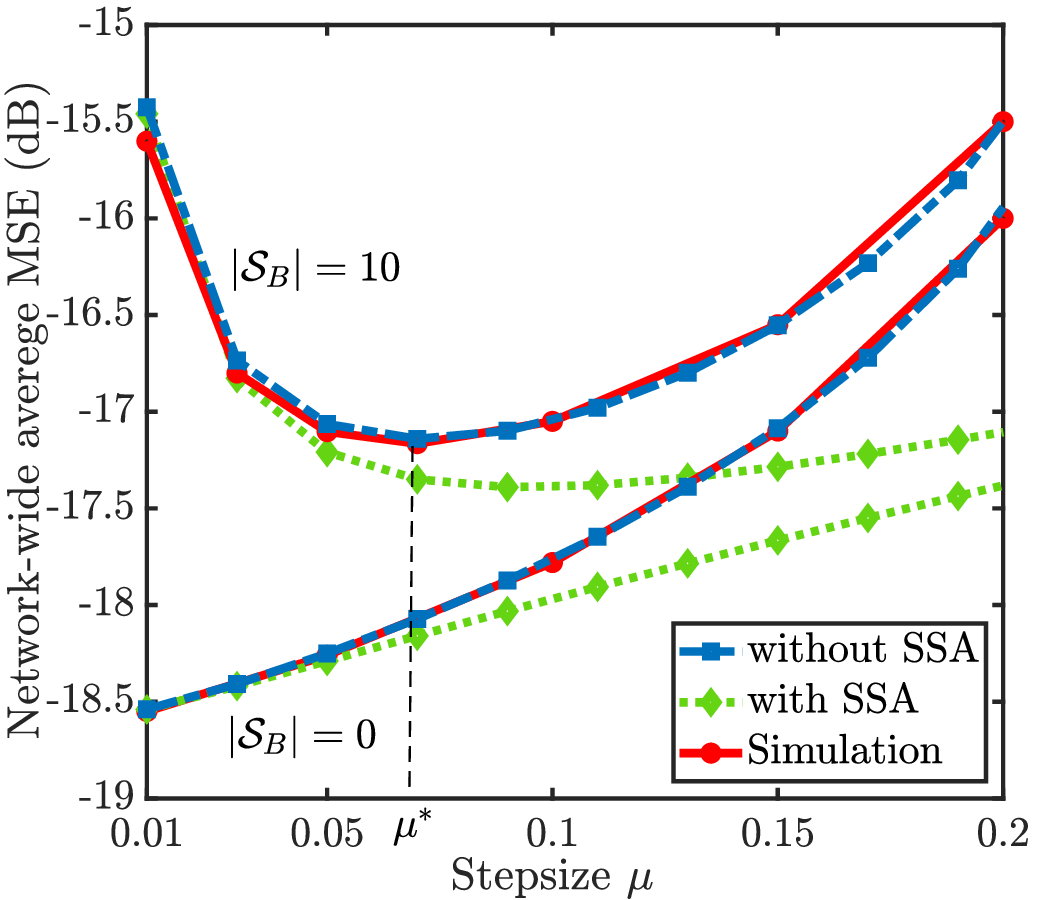}
 \caption{Effect of small stepsize approximation (SSA) on network-wide average MSE of PSO-Fed, numbers of Byzantine clients $|\Scal_B | \in \{0,10\}$, attack strength $\sigma_{B}^2 = 0.5$ and attack probability $p_a = 0.25$.}
 \label{fig:fig7}
\end{figure} 
\begin{figure}[t!]
 \centering
 \includegraphics[width=.485\textwidth]{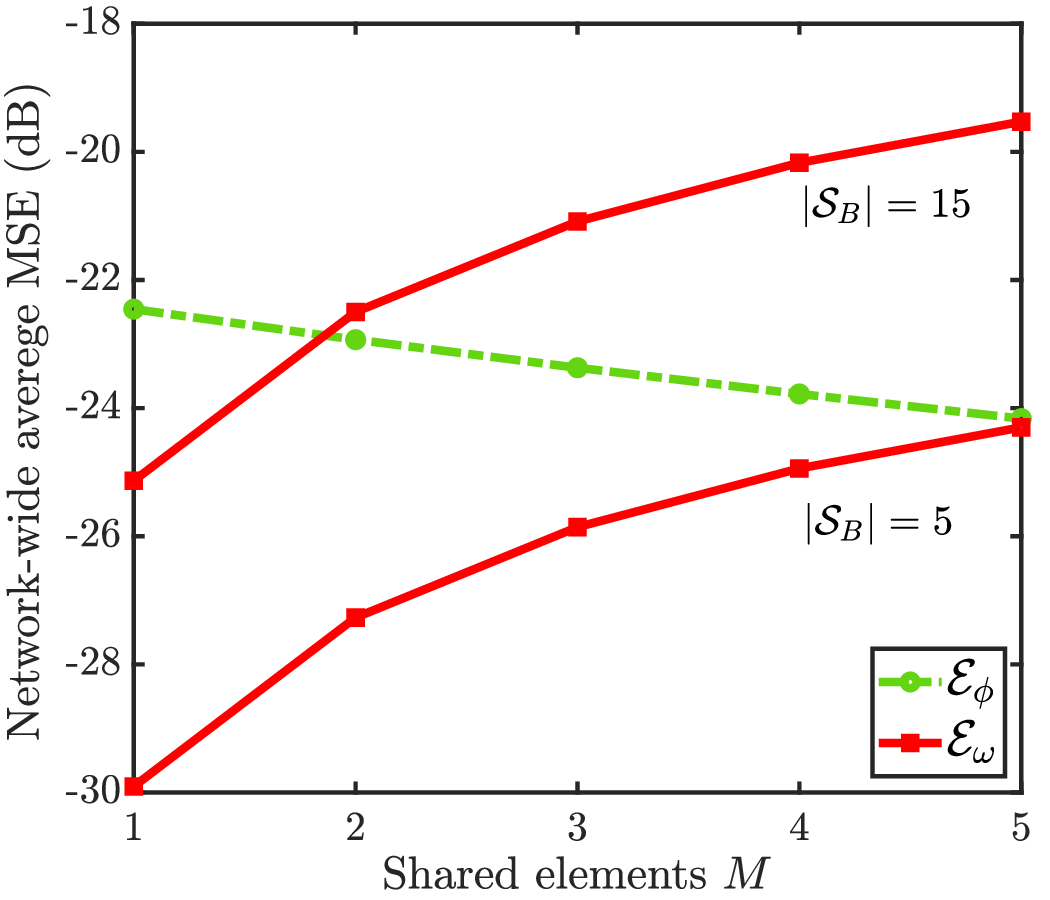}
 \caption{Effect of number of shared elements $M$ on $\Ecal_{\boldsymbol{\omega}}$ and $\Ecal_{\boldsymbol{\phi}}$ in \eqref{SSMSE} for $|\Scal_B| \in \{5,15\}$ Byzantine clients, attack strength $\sigma_{B}^2 = 0.5$ and attack probability $p_a = 0.2$.}
 \label{fig:fig8}
\end{figure} 

In our first experiment, we examine the effect of different numbers of Byzantine clients $|\Scal_B|$ on the steady-state test MSE for PSO-Fed, Online-Fed,  SignSGD, CS-Fed, and QS-Fed. In this experiment, We have $K = 100$ clients. In every iteration $n$, the server randomly selects a set of $|\Scal_n|=5$ clients to participate in FL, with each client having an equal probability of being selected. In addition, all Byzantine clients are characterized by model-poisoning noise variance of $\sigma_{B}^2 = 0.25$ and attack probability of $p_a = 1$. 
In each iteration, we allocate the same communication budget to all algorithms. Therefore, to allow all algorithms converge to their lowest steady-state MSD within $3000$ iterations, we set $\mu=0.15$ for PSO-Fed, Online-Fed, CS-Fed and QS-Fed, and $\mu=0.08$ for SignSGD at all clients. According to \eqref{rhoFcond}, the maximum stepsize ensuring the mean-square convergence of PSO-Fed in this experiment is $\mu_{max}=0.245$.
The results displayed in Fig.~\ref{fig:fig1} indicate that PSO-Fed outperforms the existing algorithms across all considered number of Byzantine clients. Notably, Online-Fed and PSO-Fed perform similarly when there is no Byzantine client or poisoning attack (i.e., $\sigma_{B}^2 = 0$). However, the performance of Online-Fed deteriorates more than PSO-Fed as the number of Byzantine clients increases. While CS-Fed performs closely to PSO-Fed in our experiment, it is important to note that the computational complexity of CS-Fed exceeds that of PSO-Fed.

In our second experiment, we examine the impact of the number of shared entries, $M$, on the steady-state test MSE of PSO-Fed. We simulate PSO-Fed with $K = 50$ clients, $D = 5$, $|\Scal_B| \in \{5,15\}$ Byzantine clients, attack strength $\sigma_{B}^2 = 0.5$, and attack probability $p_a = 0.2$. The results are shown in Fig.~\ref{fig:fig3}. We observe that increasing $M$ leads to higher steady-state test MSE. In addition, PSO-Fed exhibits greater resilience to model-poisoning attacks compared to Online-Fed, without incurring any extra computational or communication burden on the clients. This experiment corroborates our theoretical findings discussed in section \ref{SecMSE}.

In our third experiment, we investigate the effect of varying the attack strength $\sigma_{B}^2$ on the network-wide average steady state MSE. We simulate PSO-Fed with $K = 50$ clients, $D = 5$, $|\Scal_B| \in \{5,15\}$ Byzantine clients, attack strength $\sigma_{B}^2 \in \{0,0.25,0.5,0.75,1\}$, and attack probability $p_a = 0.2$. The server randomly selects $5$ clients in each iteration. The results depicted in Fig.\ref{fig:fig2} align closely with our theoretical predictions. We notice an upward trend in the network-wide average steady-state MSE as the attack strength or the number of Byzantine clients increases. In addition, the presence of $|\Scal_B| = \alpha_1$ Byzantine clients with an attack strength of $\sigma_{B}^2 = \beta_1$ results in the same MSE as having $|\Scal_B| = \alpha_2$ Byzantine clients with an attack strength $\sigma_{B}^2 = \beta_2$, provided that the condition $\alpha_1 \beta_1 = \alpha_2 \beta_2$ is met. 

In our fourth experiment, we explore how the attack probability $p_a$ influences the steady-state test MSE of PSO-Fed, considering different numbers of shared entries $M \in \{1,5\}$, with the number of Byzantine clients set to $|\Scal_B| = 5$ and attack strength to $\sigma_{B}^2 = 0.25$. The results, presented in Fig. \ref{fig:fig4_2}, align with observations from the previous experiment, showing that an increase in the attack probability leads to a higher MSE. Moreover, partial sharing (i.e., when $M < D$) enhances the resilience against model-poisoning attacks as elaborated upon in section \ref{SecMSE}.

In our fifth experiment, we investigate the impact of attack probability $p_a$ on the steady-state test MSE of PSO-Fed, considering different numbers of Byzantine clients $|\Scal_B| \in \{5,15\}$ and setting the attack strength as $\sigma_{B}^2 = 0.25$. The results shown in Fig. \ref{fig:fig4} indicate that increasing the attack probability increases the steady-state MSE. 

In our sixth experiment, we evaluate the performance of PSO-Fed using different stepsize values and Byzantine client numbers in the presence or absence of model-poisoning attacks. 
We simulate PSO-Fed with $K = 50$ clients, $D = 5$, attack strength $\sigma_{B}^2 = 0.25$, attack probability $p_a = 0.25$. The server randomly selects $5$ clients in each iteration. The results illustrated in Fig.~\ref{fig:fig5}, show that with no model-poisoning attack ($|\Scal_B| = 0$), the larger the stepsize $\mu$, the worse PSO-Fed performs, a trend consistent with gradient descent optimization methods. However, with model-poisoning attacks ($|\Scal_B| > 0$), the performance first improves with an increase in $\mu$, then deteriorates when $\mu$ grows larger. This observation confirms the existence of an optimal stepsize $\mu^*$ that ensures the best performance of PSO-Fed under model-poisoning attacks. By calculating the optimal stepsize using \eqref{muopt2}, we determine $\mu^* \approx 0.03$, which corresponds to the experimental results.

In our seventh experiment, we analyze the performance of PSO-Fed using different stepsizes in the presence or absence of model-poisoning attacks, considering different attack strengths $\sigma_{B}^2$. We simulate PSO-Fed with $K = 50$ clients, $D = 5$, $|\Scal_B| = 5$ Byzantine clients, and attack probability of $p_a = 0.25$. The server randomly selects $5$ clients in each iteration. The results, presented in Fig.~\ref{fig:fig6}, echo the observations from the previous experiment: without a poisoning attack ($\sigma_{B}^2 = 0$), PSO-Fed's performance decreases as the stepsize $\mu$ increases. Conversely, with model-poisoning attacks ($\sigma_{B}^2 > 0$), performance initially improves with an increase in $\mu$ but begins to decline as $\mu$ is further increased. This pattern confirms the existence of an optimal stepsize $\mu^*$ for scenarios with model-poisoning attacks. Using \eqref{muopt2} to estimate the optimal stepsize, we find $\mu^* \approx 0.03$, corresponding to the experiment's findings. This experiment also reinforces our conclusion following the third experiment that identical attack properties result in the same MSE.

In our eighth experiment, we investigate the impact of the commonly-adopted small stepsize approximation (SSA) on predicting the performance of PSO-Fed in the presence or absence of model-poisoning attacks. Specifically, we assume that higher orders of $\mu$ are negligible when computing ${\boldsymbol\Fcal}$ via \eqref{Fmat}. We then compare the simulated network-wide average MSE of PSO-Fed with the corresponding theoretical predictions with or without SSA. We simulate PSO-Fed with $K = 50$ clients, $D = 5$, attack probability $p_a = 0.25$, attack strength $\sigma_{B}^2 = 0.5$, and numbers of Byzantine clients $|\Scal_B| \in \{0,10\}$. The server randomly selects $5$ clients in each iteration. The results in Fig.~\ref{fig:fig7} reveal that, when $\mu$ is sufficiently small, SSA does not hinder the accurate prediction of steady-state MSE even without the knowledge of the $4$th moment of the input vector, $\Hcalbf$. However, as the stepsize increases, discrepancies arise in both scenarios, with and without model-poisoning attacks, diverging from the ideal case.

In our final experiment, we investigate the effect of partial sharing on different terms of MSE \eqref{SSMSE}, namely, $\Ecal_{\boldsymbol{\omega}}$ and $\Ecal_{\boldsymbol{\phi}}$, in the presence of model-poisoning attacks. We compute both terms using the same setup as in our second experiment. 
The results depicted in Fig.~\ref{fig:fig8} show that decreasing the number of shared parameters considerably lowers $\Ecal_{\boldsymbol{\omega}}$, significantly mitigating the impact of model-poisoning attacks. However, sharing fewer entries increases $\Ecal_{\boldsymbol{\phi}}$. Still, the benefits of partial sharing in reducing $\Ecal_{\boldsymbol{\omega}}$ are more substantial, leading to reduced MSE. Note that $\Ecal_{\boldsymbol{\phi}}$ remains unchanged across both considered cases, as it depends solely on scheduling and data noise. Finally, this experiment corroborates our theoretical findings and confirms our assertions regarding the effect of partial sharing on $\Ecal_{\boldsymbol{\omega}}$ and $\Ecal_{\boldsymbol{\phi}}$, as discussed at the end of section~\ref{SecMSE}. 

\section{\textcolor{black}{Discussion}} \label{sec:discussion}







\textcolor{black}{In this section, we discuss our key findings, their potential impact, and the associated limitations.}

\textcolor{black}{The PSO-Fed algorithm enhances communication efficiency of online FL without imposing any additional communication or computational overhead on clients. By sharing only a fraction of the model parameters, PSO-Fed effectively improves robustness against model-poisoning attacks without requiring any supplementary mechanism or increasing the computational burden of clients. This aligns with the core objective of FL research, which seeks to minimize client-side computational demands while achieving superior accuracy compared to existing algorithms, without the need for any additional mechanism required by other methods.}

\textcolor{black}{In this work, we presented a novel analysis of the resilience of PSO-Fed to model-poisoning attacks. Our approach to characterize the attack using an intermittent attack model as defined by~\eqref{eq:attack_model} in conjunction with random client scheduling is entirely original and distinct from comparable studies like \cite{10398739}, which assume full client participation. This approach ensures that our system model aligns well with the FL requirements. In addition, the analysis of mean-square convergence revealed a key insight that the error induced by the Byzantine attack and the observation noise counteract each other as a function of the stepsize, as shown in~\eqref{SSMSE}. This insight prompted our investigation into the optimal stepsize in section~\ref{OptStep}, an aspect that, to our knowledge, has not been acknowledged or analyzed in similar studies. We demonstrated that PSO-Fed not only surpasses state-of-the-art algorithms in performance, but also incurs lower computational overhead. Unlike other competitor algorithms, PSO-Fed achieves superior MSE performance without necessitating additional mechanisms or imposing any extra computational burden on clients. This aligns with the fundamental objectives of FL research and indicates that PSO-Fed is well-suited for both researchers and practitioners in real-world scenarios.}

\textcolor{black}{The use of partial sharing to enhance resilience against Byzantine attacks has not been recognized in previous similar studies. Additionally, our identification and analysis of the optimal stepsize represent an unexplored aspect in comparable studies, introducing a new dimension to this research. The closed-form expression for the optimal stepsize is highly advantageous in practical applications as it allows practitioners to fine-tune their systems for optimal performance without relying on extensive numerical simulations or costly experiments. The literature highlights its application and utility in various contexts, such as channel equalization~\cite{7078090}. Choosing the optimal stepsize not only accelerates algorithm convergence but also enhances its stability and efficiency in real-world deployments. This is particularly valuable in scenarios where computational resources and time are critical, such as in applications involving FL and IoT.} 

\textcolor{black}{To facilitate our theoretical analysis, we have made relatively mild and commonly accepted assumptions about the statistical properties of the input vectors of each client. Specifically, they are drawn from WSS multivariate random processes. This assumption enables us to leverage the analytical properties of the WSS processes and make our stability and performance analysis more tractable. Furthermore, we assume independence among the observation noise, attack signal, and selection matrices across all clients and iterations. These assumptions are not merely mathematical conveniences as they are employed in practical scenarios where such conditions are often met, such as in communication systems where noise is modeled as IID Gaussian, or in decentralized systems where each client operates independently. However, the validity of these assumptions in real-world applications must be carefully considered. In practice, deviations from stationarity or the presence of correlated noise and attack signals can occur, potentially impacting system performance. If these assumptions do not hold, the accuracy of our theoretical predictions may be affected. While our assumptions provide a strong foundation for theoretical analysis, future work can explore relaxing these assumptions to enhance the applicability of our approach to more complex and realistic scenarios.}

\textcolor{black}{Our work has been limited to a linear regression task. Nevertheless, both the methodology employed and the findings can be extended to other applications or domains, such as random Fourier features (RFF)~\cite{rff}. However, such extensions may require more intricate procedures and potentially additional assumptions. We intend to investigate these possibilities in future research. Additionally, it may be beneficial to explore other forms of adversarial attacks, such as data poisoning, label poisoning, and evasion attacks on networks, while PSO-Fed is in operation, and subsequently gauge its performance under such conditions.}

\section{Conclusion} \label{sec:conclusion}

We conducted a theoretical analysis of the recently proposed PSO-Fed algorithm to examine its resilience to model-poisoning (Byzantine) attacks imparted by partial sharing. In our analysis, we considered a linear regression task with the local objective function of each client defined as an empirical risk. We showed that, even under Byzantine attacks, PSO-Fed converges in both mean and mean-square senses, given an appropriate choice of stepsize. Notably, we showed that, in the presence of Byzantine clients, the steady-state MSE of PSO-Fed is significantly smaller than that of the Online-Fed algorithm, which does not feature partial sharing. Our theoretical analysis also uncovered the existence of a non-trivial optimal stepsize for PSO-Fed under model-poisoning attacks.
Simulation results corroborated our theoretical findings, highlighting PSO-Fed's effectiveness against Byzantine attacks and validating the accuracy of the theoretically predicted values of its steady-state MSE and optimal stepsize.

\appendices 

\section{Evaluation of matrices ${\mathbfcal{Q}}_{\mathbfcal{A}}$ and ${\mathbfcal{Q}}_{\mathbfcal{B}}$} \label{Ap2}

Let us define
\begin{align*} 
\boldsymbol {\mathcal {A}}_{n} = 
\begin{bmatrix} \mathbf {A}_{1, 1, n} &~ \mathbf {A}_{1, 2, n} &~ {\dots }&~ \mathbf {A}_{1, K+1, n} \\ \mathbf {A}_{2, 1, n} &~ \mathbf {A}_{2, 2, n} &~ {\dots }&~ \mathbf {A}_{2, K+1, n} \\ \vdots &~ \vdots &~ \ddots &~ \vdots \\ \mathbf {A}_{K+1, 1, n} &~ \mathbf {A}_{K+1, 2, n} &~ {\dots }&~ \mathbf {A}_{K+1, K+1, n} 
\end{bmatrix},
\end{align*}
where
\begin{align*} 
\mathbf {A}_{i, j, n} = 
\begin{cases} \mathbf {I}_{D}, &~ {\mathrm {if} }~ i, j=1 \\ a_{i, n} \mathbf {S}_{i, n}, &~ {\mathrm {if} } ~i=2, \ldots, K+1, j=1 \\ \mathbf {I}_{D} - a_{i, n} \mathbf {S}_{i, n}, &~ {\mathrm {if} }~ \left ({i=j}\right) \neq 1 \\ \mathbf {0} &~ {\mathrm {otherwise}}.
\end{cases}
\end{align*}
Thus, ${\mathbfcal{Q}}_{\mathbfcal{A}}$ is given by
\begin{align*}
&\mathbb {E} \left [{\boldsymbol {\mathcal {A}}_{n} \otimes _{b} \boldsymbol {\mathcal {A}}_{n} }\right]= \\
&\mathbb {E} 
\begin{bmatrix} \mathbf {A}_{1, 1, n} \otimes _{b} \boldsymbol {\mathcal {A}}_{n} &~ {\dots }&~ \mathbf {A}_{1, K+1, n} \otimes _{b} \boldsymbol {\mathcal {A}}_{n} \\ \mathbf {A}_{2, 1, n}\otimes _{b} \boldsymbol {\mathcal {A}}_{n} &~ {\dots }&~ \mathbf {A}_{2, K+1, n} \otimes _{b} \boldsymbol {\mathcal {A}}_{n} \\ \vdots &~ \ddots &~ \vdots \\ \mathbf {A}_{K+1, 1, n}\otimes _{b} \boldsymbol {\mathcal {A}}_{n} &~ {\dots }&~ \mathbf {A}_{K+1, K+1, n} \otimes _{b} \boldsymbol {\mathcal {A}}_{n} 
\end{bmatrix}
\end{align*}
where
\begin{align*}
&\mathbb {E} \left [{\mathbf {A}_{i, j, n} \otimes _{b} \boldsymbol {\mathcal {A}}_{n} }\right]= \\&\quad  \mathbb {E} \begin{bmatrix} \mathbf {A}_{i, j, n} \otimes \mathbf {A}_{1, 1, n} &~ {\dots }&~ \mathbf {A}_{i, j, n} \otimes \mathbf {A}_{1, K+1, n} \\ \mathbf {A}_{i, j, n}\otimes \mathbf {A}_{2, 1, n} &~ {\dots }&~ \mathbf {A}_{i, j, n} \otimes \mathbf {A}_{2, K+1, n} \\ \vdots &~ \ddots &~ \vdots \\ \mathbf {A}_{i,j, n}\otimes \mathbf {A}_{K+1, 1, n} &~ {\dots }&~ \mathbf {A}_{i, j, n} \otimes \mathbf {A}_{K+1, K+1, n} 
\end{bmatrix}.
\end{align*}

Recall that the probability of any model parameter being shared with the server in any iteration is $p_e = \frac{M}{D}$. In addition, the probability of any client being selected by the server in any iteration is $p_c = \frac{|\Scal_n|}{K}$. Therefore, we have 
\begin{align*}
\mathbb {E} \left[ a_{i, n} a_{i^{\prime}, n} \right] = \begin{cases} 
p_c &~ {\mathrm {if} }~ i = i^{\prime}  \\
p_c \left( \frac{|\Scal_n|-1}{K-1} \right) &~ {\mathrm {if} }~ i \neq i^{\prime}
\end{cases}
\end{align*}
and $\mathbb {E} \left[ \mathbf {S}_{i, n} \otimes \mathbf {S}_{i, n} \right]$ is a diagonal matrix with the $z$th diagonal entry being
\begin{align*}
\begin{cases} 
p_e &~ {\mathrm {if} }~ z = (n-1)D+n,~ n=1,\cdots,D  \\
p_e \left( \frac{M-1}{D-1} \right) &~ {\mathrm {if} }~ z \neq (n-1)D+n,~ n=1,\cdots,D.
\end{cases}
\end{align*}
Subsequently, the $z$th diagonal entry of the diagonal matrix $\Scalbf = \mathbb {E} \left [a_{i, n} \mathbf {S}_{i, n} \otimes a_{i^{\prime}, n} \mathbf {S}_{i, n}\right] \in \mathbb {R}^{D^2 \times D^2}$ is
\begin{align*}
\begin{cases} 
p_c p_e &~ {\mathrm {if} }~ i = i^{\prime} ~ {\mathrm {and} }~ \big(z = (n-1)D+n,\\
 & \quad ~ n=1,\cdots,D \big)\\
p_c p_e \left( \frac{M-1}{D-1} \right) &~ {\mathrm {if} }~ i = i^{\prime} ~ {\mathrm {and} }~ \big(z \neq (n-1)D+n,\\
 & \quad ~ n=1,\cdots,D \big)\\
p_c p_e \left( \frac{|\Scal_n|-1}{K-1} \right) &~ {\mathrm {if} }~ i \neq i^{\prime} ~ {\mathrm {and} }~ \big(z = (n-1)D+n,\\
 & \quad ~ n=1,\cdots,D \big)\\
p_c p_e \left( \frac{M-1}{D-1} \right) \left( \frac{|\Scal_n|-1}{K-1} \right) &~ {\mathrm {if} }~ i \neq i^{\prime} ~ {\mathrm {and} }~ \big(z \neq (n-1)D+n,\\
 & \quad ~ n=1,\cdots,D \big).
\end{cases}
\end{align*}
Therefore, we have
\begin{align*}
&\mathbb {E} \left [{\mathbf {A}_{i, j, n} \otimes \mathbf {A}_{l, m, n} }\right] =\\& \begin{cases} 
\mathbf{I}_{D^2} &~ {\mathrm {if} }~ i, j=1 ~{\mathrm {and} }~ l, m=1  \\
p_e p_c \mathbf{I}_{D^2} &~ {\mathrm {if} }~ i, j=1 ~{\mathrm {and} }~ l\geq2, m=1  \\
(1-p_e p_c) \mathbf{I}_{D^2} &~ {\mathrm {if} }~ i, j=1 ~{\mathrm {and} }~ (l= m)\neq1  \\
p_e p_c \mathbf{I}_{D^2} &~ {\mathrm {if} }~ i\geq2, j=1 ~{\mathrm {and} }~ l, m=1  \\
(1-p_e p_c) \mathbf{I}_{D^2} &~ {\mathrm {if} }~ (i=j)\neq1 ~{\mathrm {and} }~ l, m=1   \\
\Scalbf &~ {\mathrm {if} }~ (i=l)\geq2 ~{\mathrm {and} }~ (j=m)=1\\
p_e p_c \mathbf{I}_{D^2}-\Scalbf &~ {\mathrm {if} }~ i\geq2, j=1 ~{\mathrm {and} }~ (l=m)\neq 1  \\
p_e p_c \mathbf{I}_{D^2}-\Scalbf &~ {\mathrm {if} }~ (i=j)\neq 1 ~{\mathrm {and} }~ l\geq2, m=1  \\
(1-2p_e p_c) \mathbf{I}_{D^2} + \Scalbf &~ {\mathrm {if} }~ (i=j,l=m) \neq 1 \\
\mathbf{0}&~ {\mathrm {otherwise.} }
\end{cases}
\end{align*}

We similarly have
\begin{align*} 
\boldsymbol {\mathcal {B}}_{n} = 
\begin{bmatrix} \mathbf {B}_{1, 1, n} &~ \mathbf {B}_{1, 2, n} &~ {\dots }&~ \mathbf {B}_{1, K+1, n} \\ \mathbf {B}_{2, 1, n} &~ \mathbf {B}_{2, 2, n} &~ {\dots }&~ \mathbf {B}_{2, K+1, n} \\ \vdots &~ \vdots &~ \ddots &~ \vdots \\ \mathbf {B}_{K+1, 1, n} &~ \mathbf {B}_{K+1, 2, n} &~ {\dots }&~ \mathbf {B}_{K+1, K+1, n} 
\end{bmatrix},
\end{align*}
where
\begin{align*} \mathbf {B}_{i, j, n} = \begin{cases} \mathbf {I}_{D} - \sum \limits _{k=1}^{K} \frac {a_{k, n}}{|\Scal_n|} \mathbf {S}_{k, n},&~ {\mathrm {if} }~ i, j=1 \\ \frac {a_{j, n}}{|\Scal_n|} \mathbf {S}_{j, n},&~ {\mathrm {if} }~ i=1, j=2, \ldots, K+1 \\ \mathbf {I}_{D},&~ {\mathrm {if} }~ \left ({i=j}\right) \neq 1 \\ \mathbf {0}, & ~ {\mathrm {otherwise}}. \end{cases} \end{align*}
Hence, ${\mathbfcal{Q}}_{\mathbfcal{B}}$ is calculated as
\begin{align*}
&\mathbb {E} \left [{\boldsymbol {\mathcal {B}}_{n} \otimes _{b} \boldsymbol {\mathcal {B}}_{n} }\right]= \\&  \mathbb {E} 
\begin{bmatrix} \mathbf {B}_{1, 1, n} \otimes _{b} \boldsymbol {\mathcal {B}}_{n} & {\dots }& \mathbf {B}_{1, K+1, n} \otimes _{b} \boldsymbol {\mathcal {B}}_{n} \\ \mathbf {B}_{2, 1, n}\otimes _{b} \boldsymbol {\mathcal {B}}_{n} & {\dots }& \mathbf {B}_{2, K+1, n} \otimes _{b} \boldsymbol {\mathcal {B}}_{n} \\ \vdots & \ddots & \vdots \\ \mathbf {B}_{K+1, 1, n}\otimes _{b} \boldsymbol {\mathcal {B}}_{n} & {\dots }& \mathbf {B}_{K+1, K+1, n} \otimes _{b} \boldsymbol {\mathcal {B}}_{n} 
\end{bmatrix},
\end{align*}
where
\begin{align*}
&\mathbb {E} \left [{\mathbf {B}_{i, j, n} \otimes _{b} \boldsymbol {\mathcal {B}}_{n} }\right]= \\
&\mathbb {E} 
\begin{bmatrix} \mathbf {B}_{i, j, n} \otimes \mathbf {B}_{1, 1, n} & {\dots}& \mathbf {B}_{i, j, n} \otimes \mathbf {B}_{1, K+1, n} \\ \mathbf {B}_{i, j, n}\otimes \mathbf {B}_{2, 1, n} & {\dots }& \mathbf {B}_{i, j, n} \otimes \mathbf {B}_{2, K+1, n} \\ \vdots & \ddots & \vdots \\ \mathbf {B}_{i,j, n}\otimes \mathbf {B}_{K+1, 1, n} & {\dots }& \mathbf {B}_{i, j, n} \otimes _{b} \mathbf {B}_{K+1, K+1, n} 
\end{bmatrix}
\end{align*}
and
\begin{align*}
&\mathbb {E} \left [{\mathbf {B}_{i, j, n} \otimes \mathbf {B}_{l, m, n} }\right] = \\
& \begin{cases} 
(1-\frac{2 K}{|\Scal_n|} p_e p_c) \mathbf{I}_{D^2}  &~ {\mathrm {if} }~ i, j=1 ~{\mathrm {and} }~ l, m=1  \\
\quad + \frac{K}{|\Scal_n|^2} \Scalbf_{1} &  \\
\qquad + \frac{K(K-1)}{|\Scal_n|^2} \Scalbf_{2} &  \\
\frac{p_e}{K} \mathbf{I}_{D^2} - \frac{1}{|\Scal_n|^2} \Scalbf_{1} &~ {\mathrm {if} }~ i, j=1 ~{\mathrm {and} }~ l=1, m\geq2  \\
\quad - \frac{K-1}{|\Scal_n|^2} \Scalbf_{2} &  \\
\frac{p_e}{K} \mathbf{I}_{D^2} - \frac{1}{|\Scal_n|^2}  \Scalbf_{1} &~ {\mathrm {if} }~ i=1, j\geq2 ~{\mathrm {and} }~ l, m=1  \\
\quad - \frac{K-1}{|\Scal_n|^2} \Scalbf_{2} &  \\
(1-p_e) \mathbf{I}_{D^2} &~ {\mathrm {if} }~ i, j=1 ~{\mathrm {and} }~ (l= m)\neq1  \\
(1-p_e) \mathbf{I}_{D^2} &~ {\mathrm {if} }~ (i=j)\neq1 ~{\mathrm {and} }~ l, m=1   \\
\frac{1}{|\Scal_n|^2} \Scalbf &~ {\mathrm {if} }~ i=1, j\geq2  ~
{\mathrm {and} }~ l=1, m\geq2  \\
\frac{p_e}{K} \mathbf{I}_{D^2} &~ {\mathrm {if} }~ i=1, j\geq2 ~ 
{\mathrm {and} }~ (l=m)\neq1  \\
\frac{p_e}{K} \mathbf{I}_{D^2} &~ {\mathrm {if} }~ (i=j)\neq1 ~
{\mathrm {and} }~ l=1, m\geq2  \\
\mathbf{I}_{D^2} &~ {\mathrm {if} }~ (i=j)\neq1 ~{\mathrm {and} }~ (l=m)\neq1 \\
\mathbf{0}&~ {\mathrm {otherwise.} }
\end{cases}
\end{align*}
Here, $\Scalbf_{1}$ and $\Scalbf_{2}$ are diagonal matrices with their $z$th diagonal entry being
\begin{align*}
\begin{cases} 
p_c p_e &~ {\mathrm {if} }~ z = (n-1)D+n,~ n=1,\cdots,D  \\
p_c p_e \left( \frac{M-1}{D-1} \right) &~ {\mathrm {if} }~ z \neq (n-1)D+n,~ n=1,\cdots,D,
\end{cases}
\end{align*}
and
\begin{align*}
\begin{cases} 
p_c p_e \left( \frac{|\Scal_n|-1}{K-1} \right) &~ {\mathrm {if} }~ z = (n-1)D+n,\\
 & \quad ~ n=1,\cdots,D \big)\\
p_c p_e \left( \frac{M-1}{D-1} \right) \left( \frac{|\Scal_n|-1}{K-1} \right) &~ {\mathrm {if} }~ z \neq (n-1)D+n,\\
 & \quad ~ n=1,\cdots,D \big),
\end{cases}
\end{align*}
respectively. 

\section{Evaluation of matrix $\mathbfcal{H}$} \label{Ap1}

We have
\begin{align} \label{hcalbf1}
&\Hcalbf = \E[{\Xbf}_n {\Xbf}_n^\intercal \otimes_b {\Xbf}_n {\Xbf}_n^\intercal] = \\
& \E \begin{bmatrix}
    \textbf{0} & \textbf{0} &  \dots  & \textbf{0} \\
    \textbf{0} & {\xbf}_{1,n} {\xbf}_{1,n}^\intercal \otimes_b {\Xbf}_n {\Xbf}_n^\intercal & \dots  & \textbf{0} \\
    \vdots & \vdots & \ddots & \vdots \\
    \textbf{0} & \textbf{0} & \dots  &  {\xbf}_{K,n} {\xbf}_{K,n}^\intercal \otimes_b {\Xbf}_n {\Xbf}_n^\intercal  \notag
\end{bmatrix},
\end{align}
with
\begin{align} \label{hcalbf2}
&\E[{\xbf}_{1,n} {\xbf}_{1,n}^\intercal \otimes_b {\Xbf}_n {\Xbf}_n^\intercal] = \\
& \E \begin{bmatrix}
    \textbf{0} & \textbf{0} &  \dots  & \textbf{0} \\
    \textbf{0} & {\xbf}_{1,n} {\xbf}_{1,n}^\intercal \otimes {\xbf}_{1,n} {\xbf}_{1,n}^\intercal & \dots  & \textbf{0} \\
    \vdots & \vdots & \ddots & \vdots \\
    \textbf{0} & \textbf{0} & \dots  &  {\xbf}_{1,n} {\xbf}_{1,n}^\intercal \otimes {\xbf}_{K,n} {\xbf}_{K,n}^\intercal \notag
\end{bmatrix}.
\end{align}
Given that the input vectors ${\xbf}_{k,n}$ of each client arise from a wide-sense stationary multivariate random process with the covariance matrix ${\bf R}_{k} = \Ebb [{\xbf}_{k,n} {\xbf}_{k,n}^{\intercal}]$, we have
\begin{align} \label{hcalbf3}
\E[{\xbf}_{i,n} {\xbf}_{i,n}^\intercal \otimes & \ {\xbf}_{j,n} {\xbf}_{j,n}^\intercal]=\notag\\
&    \begin{cases}
    \E[{\xbf}_{k,n} {\xbf}_{k,n}^\intercal \otimes {\xbf}_{k,n} {\xbf}_{k,n}^\intercal] & \text{if $i = j = k$}\\
    {\bf R}_{i} \otimes {\bf R}_{j}  & \text{if $i \neq j$}.
    \end{cases} 
\end{align}
Each entry of $\E[{\xbf}_{k,n} {\xbf}_{k,n}^\intercal \otimes {\xbf}_{k,n} {\xbf}_{k,n}^\intercal] \in \mathbb{R}^{D \times D}$ can be computed by utilizing the Isserlis’ theorem \cite{isserlis1918formula} as 
\begin{equation} \label{hcalbf4}
    \begin{cases}
     \E[ {x}_{k,l,n}^4 ] = 3 \sigma_{ll}^4 & \forall k,l,n \\
     \E[ {x}_{k,l,n}^2 {x}_{k,m,n}^2 ]  = \sigma_{ll}^2 \sigma_{mm}^2 + 2 \sigma_{lm}^2 & \text{if $l \neq m$} \\     
     \E[ {x}_{k,l,n}^3 {x}_{k,m,n}]  = 3 \sigma_{ll}^2 \sigma_{lm}^2 & \text{if $l \neq m$} \\
     \E[ {x}_{k,l,n}^2 {x}_{k,m,n} {x}_{k,q,n}]  = \sigma_{ll}^2 \sigma_{mq}^2 + 2 \sigma_{lq}^2 \sigma_{lm}^2 & \text{if $l \neq m \neq q$}. \\
    \end{cases} 
\end{equation}
Considering that entries of ${\xbf}_{i,n}$ are independent of each other, i.e., $\sigma_{lm} = 0$, and assuming an identical variance among all entries, i.e., $\sigma_{ll} = \sigma_k$, we can simplify \eqref{hcalbf4} as
\begin{equation} \label{hcalbf5}
    \begin{cases}
     \E[ {x}_{k,l,n}^4 ] = 3 \sigma_k^4 & \forall k,l,n\\
     \E[ {x}_{k,l,n}^2 {x}_{k,m,n}^2 ]  = \sigma_k^4 & \text{if $l \neq m$} \\
     \E[ {x}_{k,l,n}^3 {x}_{k,m,n}]  = 0 & \text{if $l \neq m$} \\
     \E[ {x}_{k,l,n}^2 {x}_{k,m,n} {x}_{k,q,n}]  = 0 & \text{if $l \neq m \neq q$}. \\
    \end{cases} 
\end{equation}

\section{Evaluation of matrix ${\mathbfcal{Q}}_{\mathbfcal{C}}$} \label{ApQC}

Similar to ${\mathbfcal{Q}}_{\mathbfcal{B}}$, we can write ${\mathbfcal{Q}}_{\mathbfcal{C}}$ as
\begin{align*}
&\mathbb {E} \left [{\boldsymbol {\mathcal {C}}_{n} \otimes _{b} \boldsymbol {\mathcal {C}}_{n} }\right]= \\&  \mathbb {E} 
\begin{bmatrix} \mathbf {C}_{1, 1, n} \otimes _{b} \boldsymbol {\mathcal {C}}_{n} & {\dots }& \mathbf {C}_{1, K+1, n} \otimes _{b} \boldsymbol {\mathcal {C}}_{n} \\ \mathbf {C}_{2, 1, n}\otimes _{b} \boldsymbol {\mathcal {C}}_{n} & {\dots }& \mathbf {C}_{2, K+1, n} \otimes _{b} \boldsymbol {\mathcal {C}}_{n} \\ \vdots & \ddots & \vdots \\ \mathbf {C}_{K+1, 1, n}\otimes _{b} \boldsymbol {\mathcal {C}}_{n} & {\dots }& \mathbf {C}_{K+1, K+1, n} \otimes _{b} \boldsymbol {\mathcal {C}}_{n} 
\end{bmatrix},
\end{align*}
where
\begin{align*} 
\mathbf {C}_{i, j, n} = 
\begin{cases} 
\frac {a_{j, n}}{|\Scal_n|} \mathbf {S}_{j, n},&~ {\mathrm {if} }~ i=1, j=2, \ldots, K+1 \\ 
\mathbf {0}, & ~ {\mathrm {otherwise}}. \end{cases} \end{align*}
and
\begin{align*}
&\mathbb {E} \left [{\mathbf {C}_{i, j, n} \otimes _{b} \boldsymbol {\mathcal {C}}_{n} }\right]= \\
&\mathbb {E} 
\begin{bmatrix} \mathbf {C}_{i, j, n} \otimes \mathbf {C}_{1, 1, n} & {\dots}& \mathbf {C}_{i, j, n} \otimes \mathbf {C}_{1, K+1, n} \\ \mathbf {C}_{i, j, n}\otimes \mathbf {C}_{2, 1, n} & {\dots }& \mathbf {C}_{i, j, n} \otimes \mathbf {C}_{2, K+1, n} \\ \vdots & \ddots & \vdots \\ \mathbf {C}_{i,j, n}\otimes \mathbf {C}_{K+1, 1, n} & {\dots }& \mathbf {C}_{i, j, n} \otimes _{b} \mathbf {C}_{K+1, K+1, n} 
\end{bmatrix}.
\end{align*}
We then have
\begin{align*}
&\mathbb {E} \left [{\mathbf {C}_{i, j, n} \otimes \mathbf {C}_{l, m, n} }\right] = \\
& \begin{cases} 
\frac{1}{|\Scal_n|^2} \Scalbf &~ {\mathrm {if} }~ i=1, j\geq2  ~
{\mathrm {and} }~ l=1, m\geq2  \\
\mathbf{0}&~ {\mathrm {otherwise,} }
\end{cases}
\end{align*}
with $\Scalbf$ as calculated above.

\section{Evaluation of vector ${\pmb{\phi}}_{\pmb{\nu}}$} \label{Ap3}

Recall $\boldsymbol{\phi}_{\boldsymbol \nu} = \mathrm{bvec}\{ \E \left[ {\Xbf}_n {\boldsymbol \Theta}_{\boldsymbol \nu} {\Xbf}_n^\intercal \right] \}$. Since ${\Xbf}_n$ is not a square matrix, we cannot use the block vectorization properties. Therefore, to facilitate the calculation of $\boldsymbol{\phi}_{\boldsymbol \nu}$, we introduce the modified versions of ${\Xbf}_n$ and ${\boldsymbol \Theta}_{\boldsymbol \nu}$ as $\hat{{\Xbf}}_n = \mathrm{bdiag}\{\textbf{0}, \mathrm{diag} \{{\xbf}_{1, n}\}, \ldots, \mathrm{diag} \{{\xbf}_{K, n}\}\}$ and $\hat{\boldsymbol \Theta}_{\boldsymbol \nu} = {\boldsymbol \Theta}_{\boldsymbol \nu} \otimes \mathbf{I}_D$. Therefore, we have
\begin{equation} \label{phiv1}
\boldsymbol{\phi}_{\boldsymbol \nu} = \E[\hat{\Xbf}_n \otimes_b \hat{\Xbf}_n] \mathrm{bvec}\{ \hat{\boldsymbol \Theta}_{\boldsymbol \nu} \},
\end{equation}
where $\E[\hat{\Xbf}_n \otimes_b \hat{\Xbf}_n ]$ can be written as
\begin{align} \label{phiv2}
& \E \begin{bmatrix}
    \textbf{0} & \textbf{0} &  \dots  & \textbf{0} \\
    \textbf{0} & {\xbf}_{1,n}\otimes_b \hat{\Xbf}_n & \dots  & \textbf{0} \\
    \vdots & \vdots & \ddots & \vdots \\
    \textbf{0} & \textbf{0} & \dots  &  {\xbf}_{K,n} \otimes_b \hat{\Xbf}_n
\end{bmatrix},
\end{align}
with
\begin{align} \label{phiv3}
&\E[{\xbf}_{1,n} \otimes_b \hat{\Xbf}_n] = \notag  \\ 
& \E \begin{bmatrix}
    \textbf{0} & \textbf{0} &  \dots  & \textbf{0} \\
    \textbf{0} & {\xbf}_{1,n} \otimes {\xbf}_{1,n}  & \dots  & \textbf{0} \\
    \vdots & \vdots & \ddots & \vdots \\
    \textbf{0} & \textbf{0} & \dots  &  {\xbf}_{1,n} \otimes {\xbf}_{K,n}
\end{bmatrix}.
\end{align}
We can calculate $\E[{\xbf}_{i,n} \otimes {\xbf}_{j,n}]$ as
\begin{equation} \label{phiv4}
    \begin{cases}
    \E[{\xbf}_{k,n} \otimes {\xbf}_{k,n}] & \text{if $i = j = k$}\\
    \mathbf{0}  & \text{if $i \neq j$},
    \end{cases} 
\end{equation}
where each entry of $\E[{\xbf}_{k,n} \otimes {\xbf}_{k,n}] \in \mathbb{R}^{D \times D}$ can be computed by utilizing the Isserlis’ theorem \cite{isserlis1918formula} as 
\begin{equation} \label{phiv5}
    \begin{cases}
     \E[ {x}_{k,l,n}^2 ] = \sigma_{ll}^2 & \forall k,l,n \\
     \E[ {x}_{k,l,n} {x}_{k,m,n}]  = \sigma_{lm} & \text{if $l \neq m$}.
    \end{cases} 
\end{equation}
Given that the entries of ${\xbf}_{i,n}$ are independent of each other, i.e., $\sigma_{lm} = 0$ and assuming an identical variance among all entries of ${\xbf}_{i,n}$, i.e., $\sigma_{ll} = \sigma_k$, we can write \eqref{phiv5} as\begin{equation} \label{phiv6}
    \begin{cases}
     \E[ {x}_{k,l,n}^2 ] = \sigma_k^2 & \forall k,l,n\\
     \E[ {x}_{k,l,n} {x}_{k,m,n}]  = 0 & \text{if $l \neq m$}.
    \end{cases} 
\end{equation}

\bibliographystyle{IEEEtran}
\bibliography{refs}

\end{document}